\documentclass[lettersize,journal]{IEEEtran}
\usepackage{amsmath,amsfonts}
\usepackage{algorithmic}
\usepackage{algorithm}
\usepackage{array}
\usepackage[caption=false,font=normalsize,labelfont=sf,textfont=sf]{subfig}

\usepackage{arydshln}
\usepackage{textcomp}
\usepackage{stfloats}
\usepackage{multirow}
\usepackage{amssymb}
\usepackage{url}
\usepackage{verbatim}
\usepackage{amsmath}
\usepackage{graphicx}
\usepackage{cite}
\usepackage{color,soul}
\usepackage[colorlinks=true, allcolors=blue]{hyperref}
\newbox{\myorcidaffilbox}
\sbox{\myorcidaffilbox}{\large\includegraphics[height=4mm]{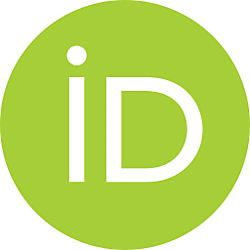}}
\newcommand{\orcidaffil}[1]{\href{https://orcid.org/#1}{\usebox{\myorcidaffilbox}}}

\begin{document}

\title{Feature Importance in Pedestrian Intention Prediction: A Context-Aware Review}

\author{Mohsen Azarmi \orcidaffil{0000-0003-0737-9204},
\and Mahdi Rezaei \orcidaffil{0000-0003-3892-421X}, 
\and He Wang \orcidaffil{0000-0002-2281-5679},
\and Ali Arabian \orcidaffil{0000-0002-6972-2584}


\thanks{M. Azarmi, M. Rezaei, and A. Arabian are with the Institute for Transport Studies, University of Leeds, LS2 9JT Leeds, U.K. (e-mail: tsmaz@leeds.ac.uk; m.rezaei@leeds.ac.uk; tsaar@leeds.ac.uk).}
\thanks{H. Wang is with the Department of Computer Science, University College London, Gower Street London, WC1E 6BT, U.K. (e-mail: he\_wang@ucl.ac.uk).}
}



\maketitle

\begin{abstract}
Recent advancements in predicting pedestrian crossing intentions for Autonomous Vehicles using Computer Vision, particularly Deep Neural Networks (DNNs) are promising. However, the black-box nature of DNNs poses challenges in understanding how the model works 
and how input features contribute to final predictions. This lack of interpretability delimits the trust in model performance 
and hinders informed decisions on feature selection, representation, and model optimisation; thereby affecting the efficacy of future research 
in the field.
To address this, we introduce Context-aware Permutation Feature Importance (CAPFI), a novel approach tailored for pedestrian intention prediction. CAPFI enables more interpretability and reliable assessments of feature importance by leveraging subdivided scenario contexts, mitigating the randomness of feature values through targeted shuffling. This aims to reduce variance and prevent biased estimations in importance scores during permutations.
We divide the Pedestrian Intention Estimation (PIE) dataset into 16 comparable context sets, measure the baseline performance of five distinct neural network architectures for intention prediction in each context, and assess input feature importance using CAPFI.
We observed nuanced differences among models across various contextual characteristics. 
The research reveals the critical role of pedestrian bounding boxes and ego-vehicle speed in predicting pedestrian intentions, and potential prediction biases due to the speed feature through cross-context permutation evaluation. We propose an alternative feature representation by considering proximity change rate for rendering dynamic pedestrian-vehicle locomotion, thereby enhancing the contributions of input features to intention prediction.
These findings underscore the importance of contextual features and their diversity to develop accurate and robust intent-predictive models. 
\\
\end{abstract}


\begin{figure}[!t]
    \centering
    \includegraphics[width=3.4in]{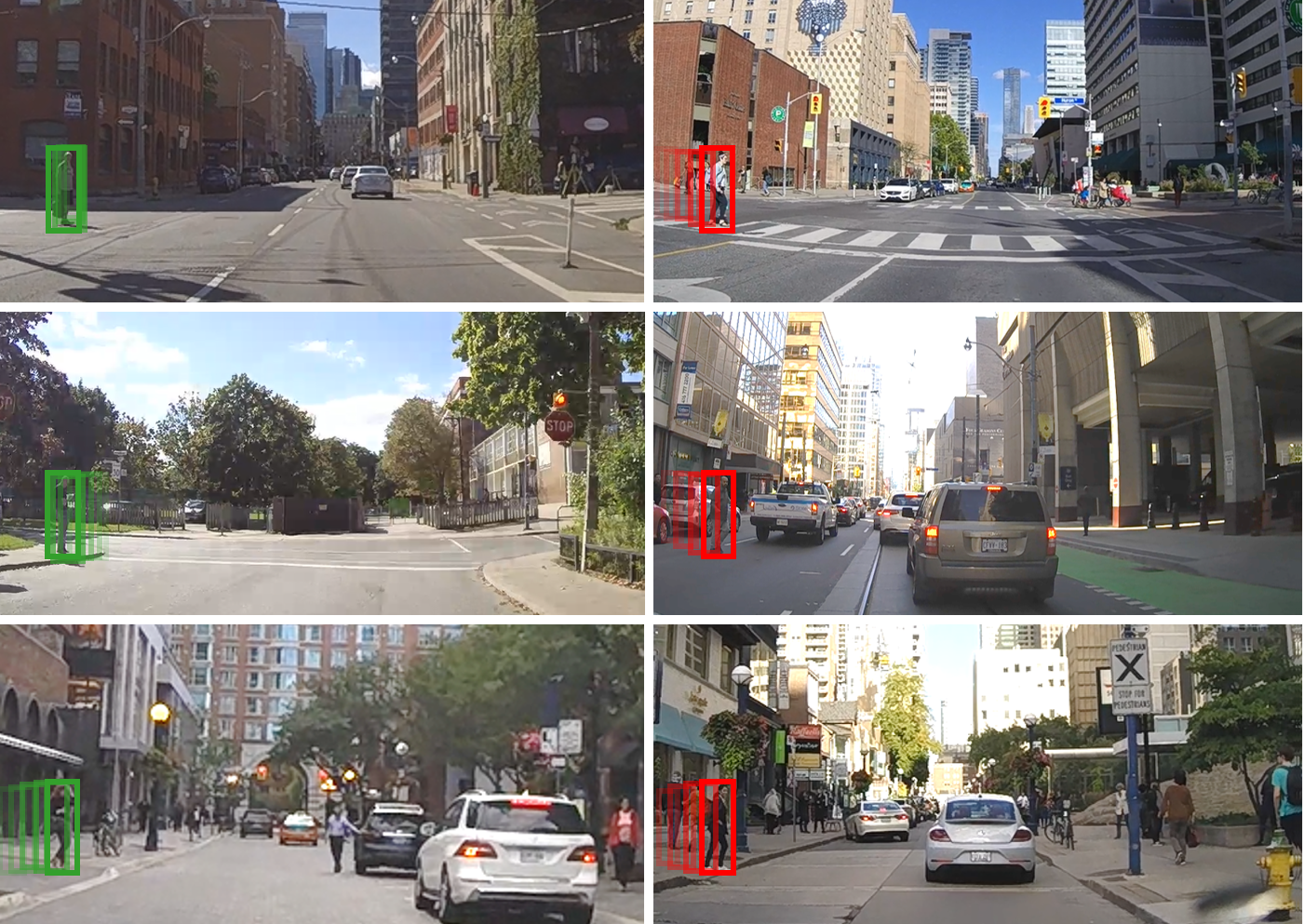}
    \caption{Pedestrian with crossing (red bounding box) and not-crossing intentions (green bounding box) in various roadway types and contexts such as 
    crosswalk designation state, traffic-light state, and also depending on the ego-vehicle speed. 
    }
\label{fig:teaser}
\end{figure}

\section{Introduction}
\IEEEPARstart{T}{he} integration of autonomous vehicles (AVs) into urban environments is a revolutionary shift in transportation, which enhances safety, efficiency, and accessibility. Central to the safe operation of AVs is their ability to accurately anticipate pedestrians' actions and respond timely, 
particularly when pedestrians 
crossing actions are likely. 
In recent years, there has been growing research interest in pedestrian intention prediction \cite{rasouli2017understanding,fang2024behavioral,galvao2023pedestrian}, thanks to the enhancement of Computer Vision techniques, particularly through learning-based methodologies like deep neural networks (DNNs), on relevant tasks including pedestrian detection 
\cite{zuo2023improving}, human action recognition\cite{pop2019multi}, trajectory and scene prediction \cite{neumann2021pedestrian}. 

Pedestrian intention prediction models typically function in two main stages. In the first stage, they extract visual cues and feature representations from sequential video images, capturing the characteristics of the pedestrian such as their moving trajectory, appearance attributes, body pose, and contextual information from the surrounding environment, including a semantic map of the entire scene \cite{kress2022pose,azarmi2023local}. In the second stage, 
a DNN model processes these extracted features by analysing their spatial and temporal dimensions using specific fusion strategies, which collaboratively contribute to 
intention prediction \cite{sharma2022pedestrian}.

While DNNs appear effective in intention prediction, their black-box nature poses challenges in understanding the contribution of each input feature to the final prediction \cite{feifel2021reevaluating}. This lack of interpretability hinders the transparency and reliability of pedestrian intention prediction systems, necessitating the development of methods to elucidate the decision-making mechanisms of these models. Moreover, this interpretability can provide insights into how the model works and aid in informed decisions on feature selection, representation, and model optimisation.

Recent studies on pedestrian intention prediction, often include ablation studies that simplify input feature sets. They train models on different combinations of feature sets and then determine which model performs better \cite{kotseruba2020they,rasouli2020PedestrianAA,kotseruba2021benchmark,Lorenzo2021CAPformerPC}. Feature removal-based technique \cite{covert2021explaining} has also been used to assess the importance of input features by disabling each feature and evaluating its impact on intent-predictive model performance \cite{azarmi2023local,azarmi2024pip}.
In this study, we conduct further experiments by randomly permuting the input feature values in the dataset instead of removing or disabling them. This concept was initially introduced by Breiman \cite{breiman2001random} to evaluate the importance of input features of random forest models. Then Fisher \textit{et al.} \cite{fisher2019all} proposed a model-agnostic version, called Permutation Feature Importance (PFI). 

Permuting feature values across random samples within the dataset preserves the input structure of the model, ensuring that the feature fusion strategy is applied consistently and the importance of each feature is evaluated in the entire feature set. However, pedestrian crossing behaviour may vary in different environmental contexts.
For instance, as illustrated in Figure \ref{fig:teaser}, the pedestrian being predicted may be at different distances from the ego-vehicle, at an intersection or midblock, with or without a crosswalk, and influenced by the traffic light's status and the ego-vehicle's speed. These factors create different safety levels; for example, a pedestrian at a well-marked crosswalk with a red traffic light for vehicles is in a higher safety level scenario compared to a pedestrian crossing midblock without a crosswalk and with fast-moving vehicles. These varying levels of safety impact the pedestrian's intention to cross the street 
\cite{ezzati2019negotiation}. Therefore, randomly permuting feature values across the entire video samples of a dataset can
result in biased estimations due to ignoring environmental context and can cause high variance estimations because of varying input feature values.

To address this, we introduce a novel approach called Context-aware PFI (CAPFI) tailored for pedestrian intention prediction. 
CAPFI enables the assessment of feature importance by evaluating the impact of permutation on model performance metrics within specified contexts. 
The main contributions of this research are highlighted as follows:

\begin{itemize}

\item \textbf{Context-aware Performance Evaluation}: We analyse the performance of five distinct neural network 
models 
with different architectures 
in predicting pedestrian intentions within 16 subsets of video samples with comparable contextual characteristics. These characteristics include roadway structure, traffic-light status, road designation state, proximity to the ego-vehicle, and the ego-vehicle speed. This analysis allows us to pinpoint risky scenario contexts and assess how well the models perform. 

\item \textbf{Permutation Feature Importance Analysis}: We conduct a comprehensive examination of the contribution of input features, including pedestrian bounding box location, body pose, local image, and vehicle speed, across five intent-predictive models. This analysis is facilitated by employing the context-aware permutation feature importance (CAPFI) technique, allowing us to gain insights into the significance of these features in various pedestrian-crossing contexts.

\item \textbf{Input Feature Representation}: We propose an alternative feature representation of the ego-vehicle locomotion by considering the pedestrian-vehicle proximity change rate. This shift in the model's focus towards pedestrian-vehicle interaction aims to reduce the potential for biased predictions influenced solely by ego-vehicle speed. 

\end{itemize}

\section{Background}\label{sec:related}
This section initially provides a broad overview of deep neural network (DNN) architectures commonly used in predictive models for pedestrian crossing intentions. We then discuss these models' input features and fusion approaches.  Table \ref{tab:pie} outlines the following subsections and the candidate models utilised in this study for feature analysis. 


\subsection{Model Architectures}
Distinct strengths in capturing complex patterns of pedestrian behaviour, environmental context, and traffic dynamics are evident in each DNN architecture.
For instance, convolutional neural networks (CNNs) excel at extracting spatial features from images, revealing visual information, such as recognising traffic users \cite{rezaei20233d}, pedestrian actions \cite{kotseruba2021benchmark}, and intentions \cite{fang2024behavioral}. While conventional CNNs, which use 2D convolution operators, may struggle with sequential data, 3D CNNs show improved performance instead in intent-predictive models \cite{kotseruba2021benchmark,yang2022predicting,azarmi2023local,azarmi2024pip,ham2023cipf}. 

Recurrent neural networks (RNNs) are effective at modelling temporal dependencies, capturing the sequential nature of pedestrian dynamic behaviours through a memory mechanism, like LSTM \cite{hochreiter1997long} and GRU\cite{chung2014empirical}, that enables them to preserve information about previous inputs, rendering them suitable for predicting intentions \cite{rasouli2019pie,kotseruba2020they,rasouli2020PedestrianAA,schorkhuber2022feature,ham2023cipf}.
Graph convolutional networks (GCNs) are adept at processing graph-structured data, enabling the modelling of complex relationships between pedestrians, vehicles, and environmental factors \cite{liu2020spatiotemporal,chen2021visual,cadena2022pedestrian,zhang2022st,sharma2023visual}. Moreover, Transformer architectures excel at capturing long-range dependencies and contextual information by leveraging self-attention mechanisms \cite{zhao2020exploring}, making them practical in large-scale datasets of complex traffic scenes for understanding pedestrian intentions \cite{Lorenzo2021CAPformerPC,achaji2022attention,zhou2023pit}. 

Hybrid architectures have also been studied to simultaneously accomplish multiple tasks for predicting pedestrian action \cite{pop2019multi}, future trajectory \cite{li2023pedestrian}, and crossing intention \cite{Yao2021CouplingIA,rasouli2022multi}. However, the information-sharing mechanisms between different tasks in these models can complicate the assessment of the features' contribution by creating complexity in the connection between input features and output predictions. 

The candidate models in this study, as indicated in Table \ref{tab:pie} by star sign, have all approached intention prediction as a singular task and framed it as a binary classification to determine whether the pedestrian is crossing in front of the AV or not.

\begin{table*}[t]
    \centering
    \caption{Common Pedestrian Crossing Intention Prediction Models. 
}
    \begin{tabular}{llcccccccccccccc}
        \noalign{\hrule height 1pt}
        \multicolumn{1}{c}{\multirow{2}{*}{\textbf{Model}}} & \multicolumn{1}{c}{\multirow{2}{*}{\textbf{Architecture}}} & \multicolumn{9}{c}{\textbf{Input Features}} & \multicolumn{2}{c}{\textbf{Fusion Strategy}} & \multicolumn{3}{c}{\textbf{Performance}} \\ \cline{3-16} 
        \multicolumn{1}{c}{} & \multicolumn{1}{c}{}              & \underline{L} & \underline{B} & \underline{P} & I & D & O & S & C & \underline{V} & \textbf{Stage} & \textbf{Type} & \textbf{Acc} & \textbf{AUC} & \textbf{F1} \\ \hline 
        SingleRNN \cite{kotseruba2020they}*       & CNN + LSTM   & \checkmark & \checkmark & \checkmark &  &  &  &  &  & \checkmark & Late & Con. & 0.81 & 0.75 & 0.64  \\
        SFRNN \cite{rasouli2020PedestrianAA}*     & CNN + GRU    & \checkmark & \checkmark & \checkmark &  &  &  &  &  & \checkmark & Hierarchical & Con. & 0.82 & 0.79 & 0.69  \\
        LGCF \cite{azarmi2023local}               & 3D CNN       & \checkmark & \checkmark & \checkmark &  &  & \checkmark &  &  &  & Middle & Con. + Att. & 0.81 & 0.80 & 0.71  \\
        PCPA \cite{kotseruba2021benchmark}*       & 3D CNN + GRU & \checkmark & \checkmark & \checkmark &  &  &  &  &  & \checkmark & Late & Con. + Att. & 0.87 & 0.86 & 0.77  \\
        PCIP \cite{yang2022predicting}            & 3D CNN + GRU & \checkmark & \checkmark & \checkmark &  &  &  & \checkmark &  & \checkmark & Hierarchical & Con. + Att. & 0.89 & 0.86 & 0.80 \\  
        MTL \cite{schorkhuber2022feature}         & CNN + LSTM   &  & \checkmark & \checkmark &  &  &   &  &  & \checkmark & Middle & Con. & 0.91 & 0.93 & 0.82  \\   
        MCIP \cite{ham2022mcip}                   & CNN + GRU    & \checkmark & \checkmark & \checkmark &  &  &  & \checkmark & \checkmark & \checkmark & Late & Con. + Att. & 0.89 & 0.87 & 0.81  \\    
        CAPformer \cite{Lorenzo2021CAPformerPC}*  & Transformer  & \checkmark & \checkmark & \checkmark &  &  &  &  &  & \checkmark & Late & Con. + Att. & 0.88 & 0.80 & 0.71  \\
        PIT \cite{zhou2023pit}                    & Transformer  &  & \checkmark & \checkmark &  &  &  &  & \checkmark & \checkmark & Middle & Average & 0.91 & 0.90 & 0.82  \\
        CIPF \cite{ham2023cipf}            & 2D \& 3D CNN + GRU  & \checkmark & \checkmark & \checkmark & \checkmark &  &  & \checkmark & \checkmark & \checkmark & Late & Con. + Att. & 0.91 & 0.89 & 0.84  \\ 
        PIP-Net \cite{azarmi2024pip}       & 2D \& 3D CNN + GRU  &  & \checkmark & \checkmark & \checkmark & \checkmark & \checkmark & \checkmark &  & \checkmark & Hierarchical & Con. + Att. & 0.91 & 0.90 & 0.84 \\
        GraphPlus \cite{cadena2022pedestrian}     & GCN          & \checkmark &  & \checkmark &  &  &  & \checkmark &  & \checkmark & Hierarchical & Con. + Att. & 0.89 & 0.90 & 0.81  \\        
        VMIGI \cite{sharma2023visual}*           & GCN          & \checkmark & \checkmark & \checkmark &  &  &  &  &  & \checkmark & Late & MLP & 0.92 & 0.91 & 0.87  \\
        \noalign{\hrule height 1pt}
    \end{tabular}
    \label{tab:pie}
    \vspace{-1mm}
    {\scriptsize
    \begin{flushleft}
        *: the candidate models in this study; 
        L: local context; 
        B: bounding box coordinates; 
        P: body pose; 
        I: local box; 
        D: distance w.r.t ego-vehicle; 
        O: optical flow; 
        S: semantic segmentation; 
        C: scene context; 
        V: ego-vehicle speed; 
        \underline{underlined feature}: the importance is investigated in this study.
    \end{flushleft}
    }
\end{table*}

\subsection{Model Input Features}
Recent research on predictive models for pedestrian crossing intentions has explored different features concerning pedestrians, environment representation, and ego-vehicle motions to depict the interaction context between autonomous vehicles and pedestrians.
For example, pedestrian bounding boxes (abbreviated as BBox) are inputted into a Transformer-based model to predict the crossing action \cite{achaji2022attention}. However, their model lacks visual and contextual information about the traffic scene. In another study, the entire traffic scene (the Scene context) is inputted into a CNN-based model to predict the crossing time \cite{pop2019multi}. However, their model suffers from limited generalisation due to a lack of detailed pedestrian-specific features and the inability to effectively handle varying environmental contexts.
To overcome this, a feature vector extracted by a CNN from the cropped image of the pedestrian (referred to as the Local box) is included in the input feature set of an RNN-based model \cite{ham2023cipf, azarmi2024pip}. 
Another approach involves using a convolutional feature vector of the squared cropped image, incorporating both the pedestrian and its surrounding environment (referred to as Local context), which has shown promise in improving prediction accuracy \cite{kotseruba2020they,rasouli2020PedestrianAA}. 
Additionally, pedestrian body joint locations extracted through a pose estimation algorithm have been included, demonstrating a positive impact on gait pattern recognition for predicting the likelihood of pedestrians crossing in front of the ego-vehicle \cite{hariyono2017detection,fang2018pedestrian,lorenzo2020rnn,fang2020intention,ma2022pedestrian,zhang2022st}.

The motion information of the ego-vehicle, such as speed and acceleration \cite{skovierova2018motion}, and optical flow analysis of the scene \cite{azarmi2023local}, have been empirically identified as a significant factor in improving intention prediction accuracy. 

A group of studies suggest considering global contextual features along with the previously mentioned local features of pedestrians \cite{singh2021multi,ham2022mcip,azarmi2023local}. The semantic segmentation, which involves pixel-level classification of the entire scene, is utilised as the model's input to provide information about environmental elements \cite{yang2022predicting,ham2023cipf}. However, this representation is often noisy, computationally expensive, and requires time-consuming post-processing \cite{azarmi2024pip}. 
Moreover, studies suggest that prediction accuracy can remain consistent by concentrating on a limited set of input features, including bounding box, body pose, local context, and vehicle speed \cite{schorkhuber2022feature}. 
This study analyses these four most commonly used input features
, ensuring a fair experimental configuration by employing them consistently across all models considered, as depicted in Figure \ref{fig:models}. This approach enables a thorough assessment of the influence of each feature across various architectures.




\subsection{Feature Fusion Strategies}
Incorporating multiple and multi-modal features in a model requires a feature fusion technique to aid the model in adapting the feature representations. Different strategies have been adopted for intent-predictive models depending on the input modality and DNN architecture. 
Models such as SingleRNN \cite{kotseruba2020they} and PCPA \cite{kotseruba2021benchmark} implement late fusion, integrating features after initial processing stages, whereas models like SFRNN \cite{rasouli2020PedestrianAA} and PCIP \cite{yang2022predicting} adopt a hierarchical fusion approach, gradually merging features at different levels of network layers. These strategies are crucial in efficiently capturing diverse information sources.
LGCF \cite{azarmi2023local} and PIT  \cite{zhou2023pit} employ middle fusion, integrating features at intermediate layers of the model architecture. 

Fusion techniques vary widely in implementation; for instance, MCIP \cite{ham2022mcip} and CIPF \cite{ham2023cipf} utilise concatenation operators, merging features directly and processing them in a single tensor. Moreover, models like PCPA \cite{kotseruba2021benchmark} and CAPformer \cite{Lorenzo2021CAPformerPC} incorporate multiple attention mechanisms, enabling dynamic feature weighting for the tensor values, while PIT \cite{zhou2023pit} employs average fusion, blending features uniformly. VMIGI \cite{sharma2023visual} aggregates features based on graph connectivity and applies additional concatenation followed by a Multi-Layer Perceptron (MLP).

Additionally, the order of features inputted into the model is investigated \cite{rasouli2020PedestrianAA,ham2023cipf}, demonstrating the robustness of the fusion strategy. Hence, our experiments assessing each feature's importance can also reflect the functionality of the fusion strategy within the wider context of the entire feature set. 
We assume that features consistently demonstrating high importance across various fusion strategies are likely robust indicators of pedestrian crossing intention, indicating their relevance irrespective of the fusion technique employed. Conversely, features whose importance fluctuates or diminishes under certain fusion strategies may highlight the interaction between feature representation and fusion methodology.

\section{Related Work}
This section presents relevant studies evaluating input feature importance in the development of pedestrian intention prediction models. We then explore different techniques for determining feature importance 
specifically within the field of deep learning.


\subsection{Feature Importance Analysis}
The importance of different visual features has been evaluated in separate intent-predictive models as each model is trained on specific input features such as the BBox, Local box, Local context, Scene context \cite{kotseruba2021benchmark}, as well as Pose and ego-vehicle Speed \cite{Lorenzo2021CAPformerPC}. The results indicate that BBox, Local context, and Speed are the most informative features for learning-based predictive models.
However, these studies primarily focus on the authors' proposed methods and do not thoroughly investigate the importance of features across different architectural designs.
In our research, we address this gap by assessing feature importance in various models, providing a more comprehensive evaluation.

Another method for analysing feature importance involves removing features and evaluating the impact on model performance. 
Studies \cite{azarmi2023local,azarmi2024pip} suggest that eliminating the Speed parameter results in the most notable drop in prediction performance.
This analysis involves deactivating input neurons responsible for processing that particular feature. However, this action potentially disturbs neurons learned about feature interactions and dependencies in the deeper network's layer.

On the other hand, randomly permuting feature values within the dataset while keeping the relevant neurons active enables the exploration of relationships between the target feature and other input features \cite{fisher2019all}.
For instance, Cai \textit{et al.} \cite{cai2024research} explores how various features influence the prediction outcomes of Multi-layer Perceptrons in predicting pedestrians' crossing intentions at signalised intersections. 
Asher \textit{et al.} employed the permutation feature importance technique to study the influence of environmental contextual factors on their model predictions in pedestrian population estimation \cite{asher2023predicting}. 
Loo \textit{et al.} examined pedestrian behavioural factors interacting with buses that lead to high-risk scenarios \cite{loo2023using}.
However, these models are designed for stationary camera setups and are not suitable for driving scenarios where the camera itself is also in motion.

\subsection{Feature Importance Techniques}
Various techniques have been developed to quantify the impact of features on model predictions \cite{molnar2020interpretable}. 
For DNN-based models, techniques like gradient-based class activation maps (CAM) \cite{selvaraju2017grad} compute the gradient of the output with respect to the input features, where high gradient values indicate features that significantly impact the model's prediction. 
However, they are suitable for simple tasks with single inputs such as image classification and object detection. 
The attention mechanism \cite{zhao2020exploring} highlights relevant parts of the input sequence for each output, with features attended to most frequently considered important for prediction. However, these mechanisms may encounter interpretability challenges, especially in complex models with multiple attention heads and layers. 

Building upon game theory concepts, Shapley Additive Explanations (SHAP) \cite{lundberg2017unified} provide a unified framework for computing feature importance by considering all possible combinations of features and their contributions to the prediction. While effective with tabular data, SHAP does not naturally handle sequential data \cite{bento2021timeshap}. Moreover, SHAP has primarily been applied to numerical or categorical features, making it challenging to extend to different data modalities. 
Although efforts have been made to broaden SHAP's applicability to a wider range of models \cite{chen2022explaining,chen2023algorithms}, limitations in compatibility with certain architectures or frameworks persist. This results in incompatibility issues with SHAP's explainer for intent-predictive models due to variations in module utilisation.

\begin{figure}[!t]
    \centering
    \includegraphics[width=2.5in]{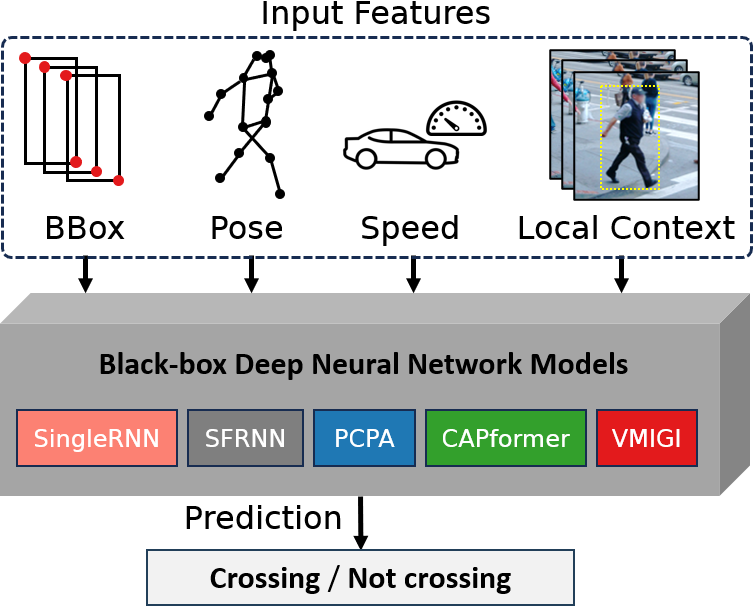}
    \caption{The candidate pedestrian intention prediction models and their input feature. These models are distinct in architecture and fusion strategy.}
\label{fig:models}
\end{figure}

\section{Methodology}
This study applies the permutation feature importance (PFI) method to evaluate the importance of each input feature in five different DNN-based model architectures for predicting pedestrian crossing intention. In contrast to traditional PFI, which randomly shuffles feature values across the entire dataset, our method, Context-aware PFI (CAPFI), shuffles values within a subset of video scenarios sharing similar contextual characteristics. 
This section initially introduces the distribution of data samples in each context. Then, we propose our alternative feature representation of ego-vehicle locomotion. Finally, the CAPFI technique used in this study to evaluate the candidate models and their input features' importance is detailed. Figure \ref{fig:models} provides an overview of the candidate models and input features that we aim to evaluate for their importance in this study. 

\subsection{Data Distribution and Subset Creation}
Pedestrian Intention Estimation (PIE) dataset \cite{rasouli2019pie}, serves as one of the largest resources for training and evaluating models to predict pedestrian crossing intention scenarios. The dataset is recorded at 30 frames per second (\textit{fps}) under daylight conditions: a sunny, clear day with high-definition (HD) resolution (1920 $\times$ 1080), spanning six hours of video capturing a total of 1841 pedestrian-vehicle interaction scenarios. 
In each interaction sample, a \textit{critical moment} is defined as the moment where both the pedestrian and the driver focus their attention on each other. All candidate models were trained to predict the pedestrian's crossing intention at this critical moment by analysing all features within the past 15 frames (0.5 seconds) before the critical moment, followed by predicting the likelihood of crossing 0.5 seconds after the critical moment. 

The dataset is originally divided into three splits 
880 video samples for training, 242 videos for validation, and 719 videos for testing. All candidate models were trained on the same samples from the training split. Hence, all evaluations in this study are conducted on a combination of test and validation samples that the models have not seen. 
However, the distribution of scenarios is imbalanced in 
contextual characteristics within the splits. 
Consequently, 
applying PFI techniques directly to this imbalanced data 
potentially leads to biased estimations and increased variance in the results due to very environmental contexts and factors.
Table \ref{tab:subsets} 
categorises the video samples into various subsets
based on pedestrian actions and contextual characteristics, according to the available annotations provided in the dataset. 
The number of samples in each set is denoted by cardinality (\textbf{C}).
A sample can belong to different subsets. For example, a video recorded at a four-way intersection with a green traffic light and the ego-vehicle accelerating exists in $S_{FW}$, $S_{Green}$, and $S_{Acc}$ subsets.

\begin{table}[htbp]
\centering
\caption{Subsets Definitions, Notations, and The Cardinality (C)}
\label{tab:subsets}
    \begin{tabular}{l@{\hskip 0.5em}|l@{\hskip 0.3em}|l@{\hskip 0.5em}|c@{\hskip 0.5em}c}
        \noalign{\hrule height 1pt}
        \textbf{Group Name} & \textbf{Scenario Context} & \textbf{Notation} & \textbf{C} \\ \hline
        \multirow{2}{*}{Crossing State} & Cross & \(S_{C}\) & {258}\\ 
         & Not Cross & \(S_{CN}\) & {634}\\ \hline
        \multirow{3}{*}{Roadway Type} & Four-Way Intersection & \(S_{FW}\) & {441}\\ 
         & Midblock Crossing & \(S_{MB}\) & {164}\\ 
         & T-Junction & \(S_{TJ}\) & {103}\\ \hline
        \multirow{3}{*}{Traffic-Light State} & Red & \(S_{Red}\) & {93}\\ 
         & Yellow & \(S_{Yellow}\) & {37}\\ 
         & Green & \(S_{Green}\) & {242}\\ \hline
        \multirow{2}{*}{Crosswalk State} & Zebra Crossing & \(S_{ZC}\) & {239}\\ 
         & Non-Zebra Crossing & \(S_{NZC}\) & {653}\\ \hline
        \multirow{3}{*}{Proximity Level} & Close Proximity & \(S_{CP}\) & {59}\\ 
         & Medium Proximity & \(S_{MP}\) & {542}\\ 
         & Far Proximity & \(S_{FP}\) & {291}\\ \hline
        \multirow{4}{*}{Ego-Vehicle Speed} & Accelerating & \(S_{Acc}\) & {216}\\ 
         & Constant & \(S_{Const}\) & {298}\\ 
         & Stopped & \(S_{Stopped}\) & {185}\\ 
         & Decelerating & \(S_{Dec}\) & {193}\\ 
         \noalign{\hrule height 1pt}
    \end{tabular}
\end{table}

\subsubsection{\textbf{Crossing State}}
This group consists of the entire 
test and validation samples from the dataset and 
the environmental context is not separated. 
This group is subcategorised by the intention label. Each video sample has a duration of 1 second, beginning 0.5 seconds before the critical moment and ending 0.5 seconds after the critical moment. 

\subsubsection{\textbf{Roadway Type}}

The potential variation in pedestrian behaviours, influenced by roadway type \cite{ezzati2019negotiation}, is captured in this categorisation. By assessing the performance of intention prediction models against various roadway types, we can uncover the strengths and weaknesses of each model specific to the given roadway, and potentially reveal unknown risks that each model may pose to pedestrians.

\subsubsection{\textbf{Traffic-Light State}} 
Subsets are formed from video samples captured at four-way intersections and T-junctions with traffic lights. 
Different traffic-light states impose varying levels of constraint or permission for pedestrian crossings, influencing the pedestrian decision-making process and the likelihood of crossing \cite{upreti2023traffic}. The prediction performance assessments provide insights into how these models respond to differences in signalisation.

\subsubsection{\textbf{Crosswalk State}} 
Crosswalk-designated areas typically offer enhanced safety and visibility for pedestrians, potentially affecting their crossing intentions compared to scenarios with no such infrastructure \cite{amado2020pedestrian}.
This differentiation is essential for assessing how accurately intent-predictive models capture the influence of designated infrastructure on pedestrian behaviour.

\subsubsection{\textbf{Proximity Level}} 
\begin{figure}[!t]
    \centering
    \includegraphics[width=3.4in]{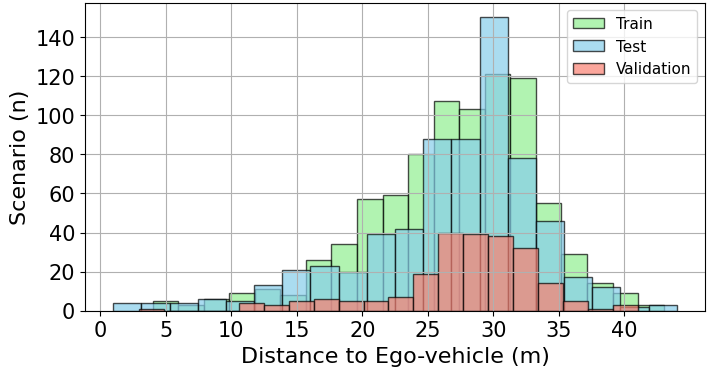}
    \caption{Histogram of proximity level of pedestrians in PIE dataset.}
\label{fig:distance}
\end{figure}

Different distance ranges may correspond to varying levels of perceived safety or risk for pedestrians, influencing their decision to cross or wait \cite{kalantari2023goes}. These subsets allow us to investigate models' performance across different ranges of distances from pedestrian to ego-vehicle. As the dataset doesn't include the distance parameter, we estimate the distances through a monocular depth estimation algorithm \cite{watson2021temporal}.
The distribution of pedestrian distances for different splits of the dataset is shown in Figure \ref{fig:distance}.
Studies show that the intent-predictive models exhibit high stability and accuracy when the longitudinal relative distance between pedestrians and the ego vehicle is approximately less than 25$m$ \cite{ma2022pedestrian}.
In our evaluation,
pedestrians located up to 15m from the ego vehicle are considered to be in close proximity; those between 15$m$ and 30$m$ are in middle proximity, and those farther than 30$m$ are in far proximity. This classification ensures a fairly balanced distribution of samples across each subset and allows for a more granular analysis of predictive performance within these ranges.

\subsubsection{\textbf{Ego-Vehicle Speed}} 
Variations in ego-vehicle speed can alter the perceived risk and urgency of crossing, thus influencing pedestrian intention \cite{zhang2023cross}. Performance analysis through this group allows us to elucidate how intent-predictive models adapt to changes in vehicular motion and the predictive factors that drive pedestrian behaviour in such scenarios. 

\subsection{Ego-vehicle Locomotion Representation}\label{sec:motion}
\begin{figure}[!t]
    \centering
    \includegraphics[width=3.4in]{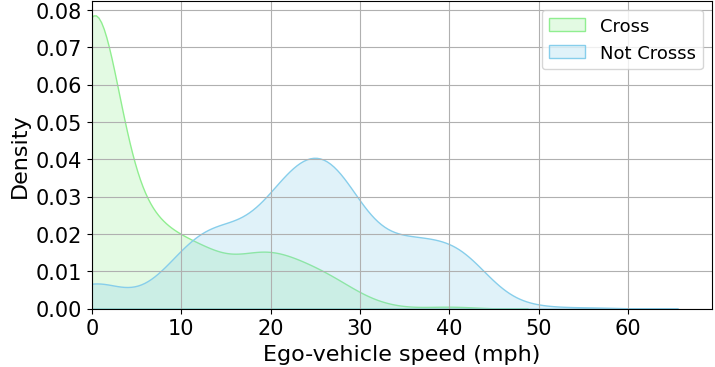}
    \caption{The Probability Distribution Function (PDF) of the ego-vehicle speed in the PIE dataset.}
\label{fig:pdf_speed}
\end{figure}

As depicted in Figure \ref{fig:pdf_speed}, most crossing samples occurred when the ego vehicle was either stationary or moving at a low speed. This prompts the question of whether models should prioritise this feature or not.
A model trained only on speed value achieves an AUC of 0.83$_{\pm0.002}$ and F1 score 0.74$_{\pm0.003}$.
Every input feature combination when it is included
, F1 score increases by over 25\% on average. It appears this model ends up learning the behaviour of the ego-vehicle driver rather than learning to predict the behaviour of pedestrians \cite{Lorenzo2021CAPformerPC}. 
Hence, we propose a proximity change rate to make this feature implicit by incorporating the rate of change in the distance between the pedestrian and the ego vehicle over time as defined as follows.

\begin{equation}\label{eq:rate_of_change}
    \Delta_{P} = \frac{\delta_{t_0} - \delta_{t_n}}{dt} 
\end{equation}
\noindent where $\Delta_{P}$ is the change in distance per meter, $\delta_{t_0}$ is the distance of the pedestrian and ego-vehicle at time $t_0$, $\delta_{t_n}$ is the distance at time $t_n$, and $dt$ is the time interval in \textit{fps} between $t_0$ and $t_n$.

In this feature representation, the model would learn to capture the dynamics of pedestrian-vehicle interaction without being directly informed about the speed of the ego-vehicle. 

\subsection{Permutation Feature Importance}
The permutation importance ($\text{PI}_{X_i}^f$) for feature $X_i$ in a given predictive model $f$ can be calculated as follows: 

\begin{equation}\label{eq:permu}
    \text{PI}_{X_i}^{f} = \frac{1}{N} \sum_{j=1}^{N} \left( \text{Metric}_{\text{baseline}} - \text{Metric}_{\text{permuted}}^{(j)} \right)
\end{equation}

\noindent where $N$ is the number of permutations for the $i$-th feature, $\text{Metric}_{\text{baseline}}$ represents the baseline performance metric (e.g., accuracy, F1 score, AUC) of the model on the dataset, $\text{Metric}_{\text{permuted}}$ is the evaluation performance of the model on the dataset obtained by permuting the feature $X_i$ in scenario context $j$.
The permuted feature is randomly shuffled across the samples while keeping the target labels fixed, effectively breaking the relationship between the particular feature and other input features. 
The permutation has been repeated for all samples included in each scenario context set ($N=$ \textbf{C}), and the shuffling pattern (random seed) for all models considered is the same. 
A higher positive value of $\text{PI}_{X_i}^f$ indicates that the feature $X_i$ is important for model $f$, as shuffling its values led to a significant decrease in performance. Conversely, a lower or negative value suggests the feature is less important.

\section{Experiments}
This section presents the evaluation of candidate models within the defined subsets of the PIE dataset's test and validation samples, considering specific contextual characteristics to measure the baseline performance for each model. Subsequently, we shift our focus to hazardous pedestrian-crossing scenarios. Following this, we evaluate the importance of input features using CAPFI across different scenario contexts. Finally, we assess the contribution of the proposed feature representation to the models' performance.

\begin{figure*}[!t]
    \centering
    \includegraphics[width=7in]{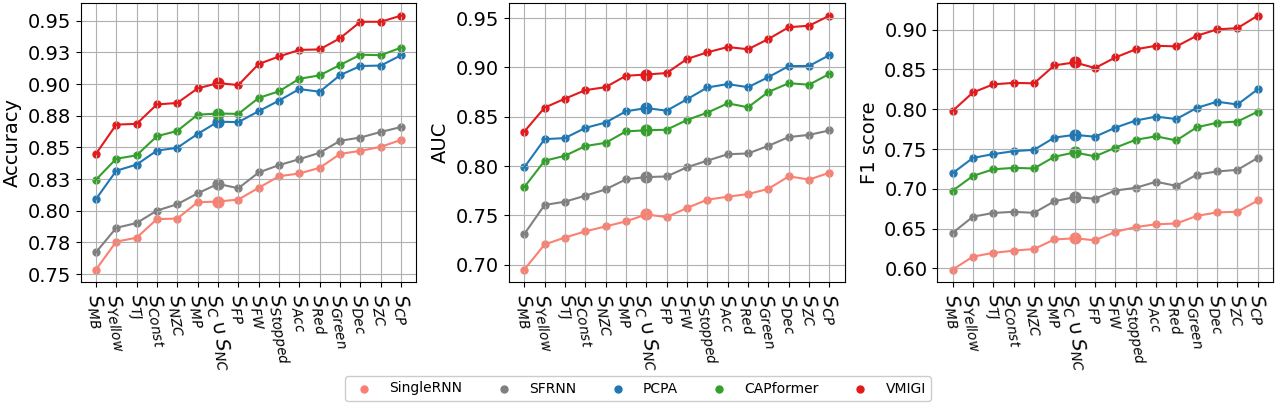}
    \caption{The performance of intention prediction models in distinct scenario contexts. The $S_C \cup S_{NC}$ displays the performance of models across all the crossing and not-crossing samples in the PIE dataset. 
    }
    \label{fig:context_perfm}
\end{figure*}

\subsection{Performance Evaluation}
The performance of intent-predictive models is evaluated using standard machine learning metrics. These metrics include \textit{Accuracy} (Acc), which quantifies the model's ability to accurately predict the binary classification of a pedestrian's intention to cross or not. However, accuracy alone may not be sufficient when the dataset is imbalanced, as it could be high even if the model fails to detect instances of a particular class (e.g., crossing intention). The \textit{area under the ROC curve} (AUC) indicates the model's proficiency in distinguishing between 
two classes of `'crossing'' or `'not crossing''. A high AUC implies that the model can effectively prioritise instances with higher probabilities of crossing. The \textit{F1 score} represents the harmonic mean of \textit{precision} and \textit{recall rate}. A high F1 score indicates that the model is effectively minimising both false positives (predicting a pedestrian intends to cross when they don't) and false negatives (failing to predict when a pedestrian intends to cross), thus contributing to pedestrian safety by reducing both types of errors.

All evaluations in this study were conducted 
on a Windows PC equipped with the Nvidia Quadro RTX A6000 GPU, an Intel Core i9 13900K 24-core processor, and 64GB of RAM. 

Figure \ref{fig:context_perfm} illustrates the re-evaluated performance of the candidate models within different scenario contexts (as per Table \ref{tab:subsets}). Overall, VMIGI surpasses all models in all contexts using GCN architecture, in terms of accuracy, AUC, and F1 Score, particularly with a significant improvement of 8.3\% in F1 score metric. 
When compared to PCAP, although CAPformer achieves higher accuracy, it exhibits lower performance in other metrics.
From another perspective, the similarity in models' performance trends across different contexts indicates the presence of challenging samples in the dataset, which almost all models struggle to predict effectively.

\subsection{High-Risk Crossing Scenarios}
\begin{figure}[!t]
    \centering
    \subfloat[]{\includegraphics[width=1.69in]{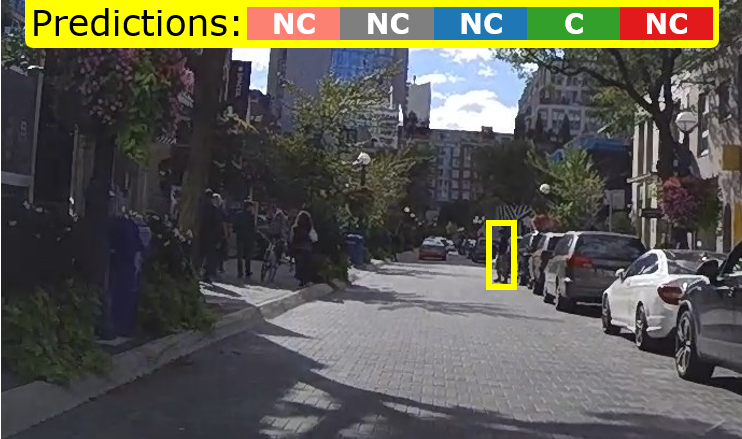}\label{subfig:hs1}}
    \hfil
    \subfloat[]{\includegraphics[width=1.69in]{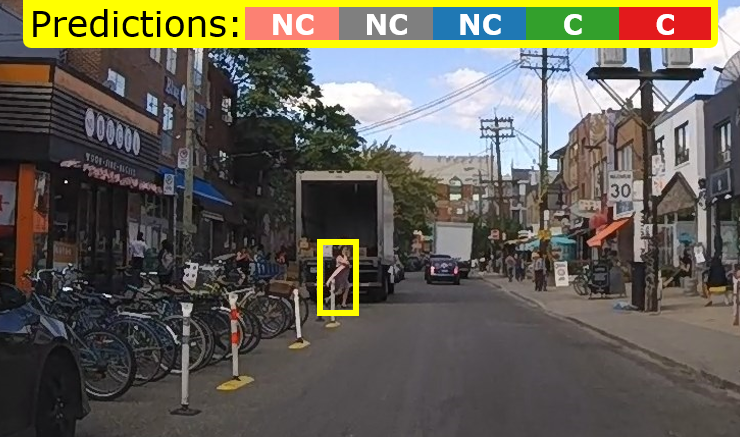}\label{subfig:hs2}}
    \hfil
    \subfloat[]{\includegraphics[width=1.69in]{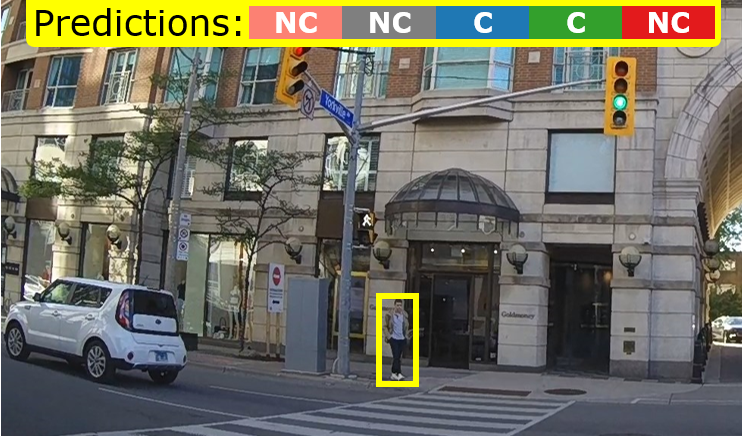}\label{subfig:hs3}}
    \hfil
    \subfloat[]{\includegraphics[width=1.69in]{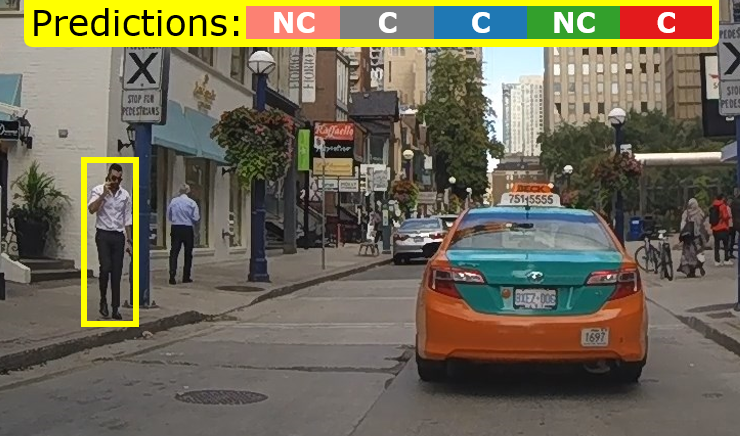}\label{subfig:hs4}}
    \hfil
    \caption{A sample for each hazardous scenario, as defined in Table \ref{tab:risky}, when all the pedestrians will cross in front of the ego-vehicle. The candidate models' predictions for crossing (C) or not-crossing (NC) intention are indicated by colour codes as defined in Figure \ref{fig:models}.}
    \label{fig:hazsample}
\end{figure}

We identify combinations of subsets that result in high-risk scenarios for pedestrians if the intent-predictive model underperforms 
to detect their intentions correctly with a very low Accuracy (Acc) and F1 score.
For instance, in $S_C \cap S_{Acc}$ scenario, the context is when the pedestrian intends to cross and the vehicle is accelerating towards it. 
Another instance (scenario C) involves  $S_C \cap S_{MB} \cap S_{NZC} \cap S_{Const}$, where the pedestrian intends to cross at a non-designated midblock, and the ego vehicle has not changed or decreased its speed.
Table \ref{tab:risky} represents the performance of the five candidate models in different scenario contexts. 
These contexts exemplify potential hazards in real-world circumstances. 
Figure \ref{fig:hazsample} depicts one corresponding sample of each hazardous scenario and the predictions have been made by the candidate models regarding pedestrian intention.

\begin{table}[htbp]
\centering
\caption{Performance Evaluation on Hazardous Scenarios}
\label{tab:risky}
\begin{tabular}{c|l@{\hskip 0.3em}|c|c@{\hskip 0.3em} c@{\hskip 0.3em}c@{\hskip 0.3em}}
\noalign{\hrule height 1pt}
\multirow{1}{*}{\textbf{\#}}  & \multicolumn{1}{c|}{\multirow{1}{*}{\textbf{Scenario Context}}}  & \multicolumn{1}{c|}{\multirow{1}{*}{\textbf{C}}}  & \multicolumn{1}{c|}{\multirow{1}{*}{\textbf{Model}}} & \multicolumn{1}{c}{\textbf{Acc}} & \multicolumn{1}{c}{\textbf{F1}} \\ \hline

\multirow{5}{*}{(a)} &\multirow{5}{*}{$S_C \cap S_{Acc}$} & \multirow{5}{*}{17} & 
   \multicolumn{1}{l|}{SingleRNN}      & 0.12 & 0.21 \\ 
&&& \multicolumn{1}{l|}{SFRNN}         & 0.29 & 0.45  \\
&&& \multicolumn{1}{l|}{PCPA}          & 0.41 & 0.58  \\ 
&&& \multicolumn{1}{l|}{CAPformer}     & 0.29 & 0.45  \\
&&& \multicolumn{1}{l|}{VMIGI}         & 0.29 & 0.45  \\ \hline
\multirow{5}{*}{(b)} &\multirow{5}{*}{$S_C \cap S_{CP} \cap S_{MB}$} & \multirow{5}{*}{6}& 
    \multicolumn{1}{l|}{SingleRNN}     & 0.17 & 0.29   \\ 
&&& \multicolumn{1}{l|}{SFRNN}         & 0.33 & 0.50  \\ 
&&& \multicolumn{1}{l|}{PCPA}          & 0.33 & 0.50  \\ 
&&& \multicolumn{1}{l|}{CAPformer}     & 0.33 & 0.50  \\ 
&&& \multicolumn{1}{l|}{VMIGI}         & 0.50 & 0.67  \\ \hline
\multirow{5}{*}{(c)} &\multirow{5}{*}{$S_C \cap S_{Green}$}& \multirow{5}{*}{28} & 
    \multicolumn{1}{l|}{SingleRNN}     & 0.21 & 0.36   \\ 
&&& \multicolumn{1}{l|}{SFRNN}         & 0.29 & 0.44  \\ 
&&& \multicolumn{1}{l|}{PCPA}          & 0.39 & 0.52  \\   
&&& \multicolumn{1}{l|}{CAPformer}     & 0.46 & 0.63  \\
&&& \multicolumn{1}{l|}{VMIGI}         & 0.43 & 0.56  \\ \hline
\multirow{5}{*}{(d)} &\multirow{5}{*}{$S_C \cap S_{MB} \cap S_{NZC} \cap S_{Const}$}& \multirow{5}{*}{8}&
    \multicolumn{1}{l|}{SingleRNN}     & 0.25 & 0.40  \\ 
&&& \multicolumn{1}{l|}{SFRNN}         & 0.25 & 0.40  \\  
&&& \multicolumn{1}{l|}{PCPA}          & 0.38 & 0.55  \\   
&&& \multicolumn{1}{l|}{CAPformer}     & 0.38 & 0.55  \\  
&&& \multicolumn{1}{l|}{VMIGI}         & 0.38 & 0.55  \\ \hline
\noalign{\hrule height 1pt}
\end{tabular}
\end{table}

The performance of the models varies depending on the scenario, with certain models exhibiting better performance in specific contexts than others. In scenarios (a) and (d), the PCPA model consistently outperforms better than other models. For scenario (b), the VMIGI model achieves the highest F1 score, and the CAPformer model demonstrates the highest score in scenario (c), indicating superior performance in that context. It's noteworthy that all samples in scenario (c) occurred when the traffic light was green for the ego-vehicle as it made the turn to the right or left road, where the pedestrian was crossing that road (see Figure \ref{subfig:hs3}). This is recognised as one of the dangerous traffic scenarios between vehicles and pedestrians \cite{matsui2023characteristics}. Therefore, considering the ego-vehicle head angle parameter may enhance the performance of models in such scenarios.

\subsection{Analytical Review on Feature Importance}

\begin{figure}[!t]
    \centering
    \label{subfig:models_pi_all}
    \includegraphics[width=3.4in]{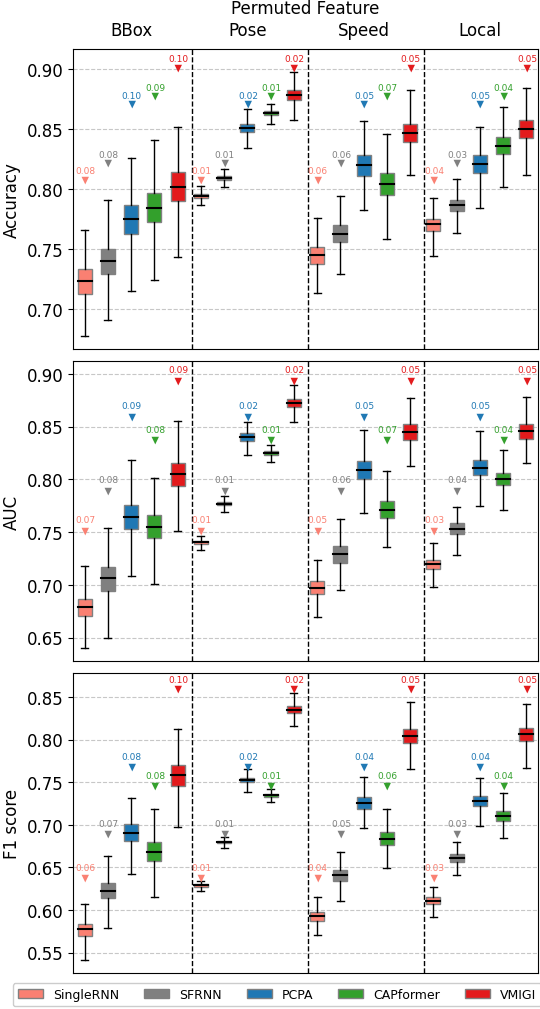}
    \caption{The performance of the candidate models after permuting each input feature. The triangle represents the baseline performance of each model, with the number above indicating the mean importance score. Features with higher scores contribute more to the model's prediction performance, as shuffling their values results in a greater decline from the baseline performance. A brief look at the graphs reveals the \textit{'BBox'} has the most important feature importance, followed by \textit{Speed'} and \textit{'Local features'} and interestingly the \textit{'Pose'} shows the lowest feature importance in predicting the pedestrian intention.}
\label{fig:models_cpfi}
\end{figure}

We initially calculated the permutation feature importance scores for the candidate models, 
using the entire test and validation samples of the dataset (including both crossing and not-crossing cases, $S_{C} \cup S_{NC}$). See Figure \ref{fig:models_cpfi} for more details. The figure illustrates how permutation affects the models' performance metrics. The input features — pedestrian BBox, Pose, Speed, and Local context — are represented as columns in the figure. Box plots within each column show the feature importance scores for the candidate models, colour-coded for clarity. Each box plot displays the interquartile range (IQR) of the permutation scores for a specific feature, spanning from the 25$^{th}$ percentile (Q1) to the 75$^{th}$ percentile (Q3). This range indicates how much the importance scores vary across repetitions ($N$). A taller box suggests greater variability (higher standard deviation, $\sigma$) and reflects more fluctuation in the model's performance when that feature's values are shuffled. The line inside each box represents the median importance score, showing the central tendency of the scores. The feature has a higher importance score if the median line is towards the bottom of the box.

\begin{figure*}[!t]
    \centering
    \includegraphics[width=7in]{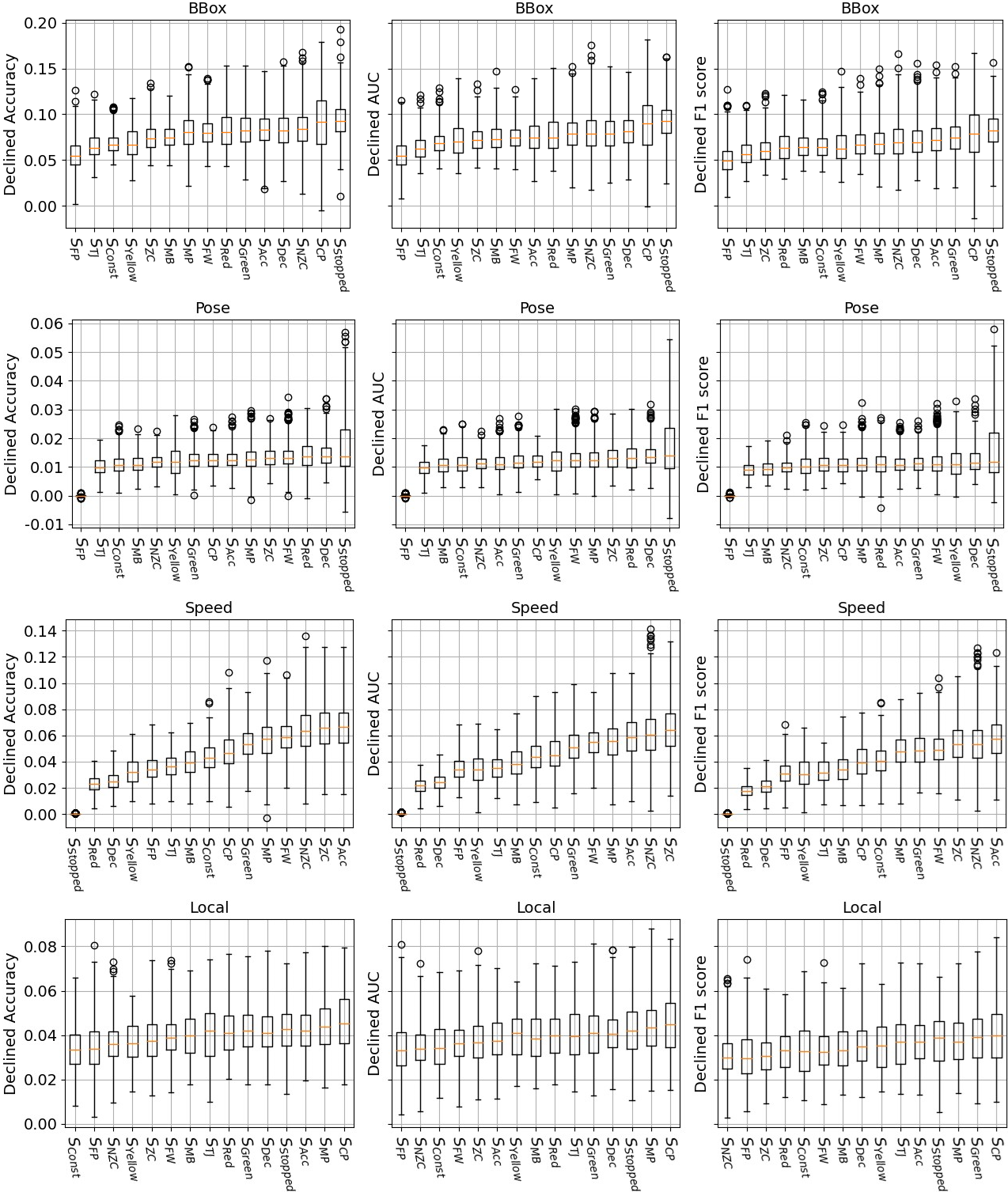}
    \caption{The result of permutation feature importance across different scenario contexts. The scenarios are sorted based on the mean importance scores on the y-axis. Feature permutation is conducted on four input features of the candidate models, assessing the performance decline rate for each feature, and repeated for the number of samples in each context ($N= \mbox{\textbf{C}}$). Features with higher importance scores contribute more to the models' prediction performance in each context.}
    \label{fig:cfpi_all}
\end{figure*}

To elucidate the importance of input features in predicting pedestrian crossing intention across various scenario contexts, CAPFI is evaluated for all the candidate models, taking into account the baseline performance metric of each model in the specific context (see Figure \ref{fig:context_perfm}). Figure \ref{fig:cfpi_all} depicts the importance of features in different scenario contexts using the proposed CAPFI technique. 

To provide a deeper insight into the conducted permutation feature importance, we analyse the results from the following aspects: 

\subsubsection{\textbf{Resemblances}}
    Despite the differences in architecture and fusion strategy used in the candidate models, their response to feature permutation has shown a striking resemblance. 
For instance, BBox is consistently more important than other features for all models, and the Local context feature maintains consistent importance across a wide range of scenario contexts.
This consistency in feature importance probably highlights the relevance of these features to the task of pedestrian crossing intention prediction, regardless of the model architecture.
    
    Furthermore, across all features for all models and almost all scenario contexts, we observed that the impact of feature permutation is most reflected in accuracy, followed by AUC, and lastly F1 score. This indicates that permutation mostly influences the models' ability to identify the crossing intentions 
    correctly.

\subsubsection{\textbf{Feature Contribution}}
    The BBox feature contributes 9.1\%$_{\sigma= 1.22}$ to accuracy, 9.2\%$_{\sigma= 1.2}$ to AUC, and 9.1\%$_{\sigma= 1.23}$ to the F1 score, achieving the highest importance scores across all models and scenarios. Conversely, the Pose feature is identified as the least important across all models and scenarios, contributing 1.3\%$_{\sigma= 0.46}$ to accuracy, 1.4\%$_{\sigma= 0.47}$ to AUC, and 1.3\%$_{\sigma= 0.46}$ to F1 score. The Speed of the ego-vehicle emerges as the second most important feature, contributing 5.1\%$_{\sigma= 2.11}$ to accuracy, 5\%$_{\sigma= 2.06}$ to AUC, and 5.1\%$_{\sigma= 2.1}$ to F1 score. While the Local context feature generally plays a supportive role in enhancing prediction performance, it is less important than BBox and often Speed. With slight variations across scenario contexts, it has contributed 4.7\%$_{\sigma= 0.71}$ to accuracy, 4.6\%$_{\sigma= 0.73}$ to AUC, and 4.7\%$_{\sigma= 0.76}$ to F1 score.

\subsubsection{\textbf{Importance Variability}}
    The BBox feature displays the greatest variability in importance scores, leading to the longest box plots among other features. The variability of BBox notably increases for video samples involving pedestrian crossings, while the feature's value changes with samples where the pedestrian is stationary. Alternatively, in samples involving pedestrian non-crossings, the feature's value changes with samples where the pedestrian is moving or the ego-vehicle is turning.
    
    The variability of Pose is the lowest among other features, indicating that body posture variations have less influence on models' prediction performance. Alternatively, this feature does not provide as much discriminative information for the pedestrian crossing intention prediction task as other features.
    This finding contradicts several studies which suggest that pedestrian pose is very important for improving performance \cite{hariyono2017detection,fang2018pedestrian,lorenzo2020rnn,fang2020intention,ma2022pedestrian,zhang2022st}.
    
    The ego-vehicle Speed variations seem to correlate with pedestrian crossing intentions in the dataset (see Figure \ref{fig:pdf_speed}), as higher vehicle speeds might be associated with scenarios where pedestrians are less likely to cross. In comparison, lower speeds might indicate situations where pedestrians are more likely to cross. Hence, exchanging the feature values between these scenarios has led to a higher variability of Speed values.

    Variability in the Local context feature's importance arises from differences in pedestrian appearance and environmental conditions. For instance, samples with varied pedestrian occlusion levels, lighting conditions, or infrastructure layouts can elicit different responses from the models, leading to fluctuations in the importance of the Local context feature.

\subsubsection{\textbf{Impact of Context Changes}}
    The importance score of BBox varies depending on different scenario contexts.
    For example, in the T-junctions scenario ($S_{TJ}$), the importance score is relatively lower compared to both four-way intersections ($S_{FW}$) and midblock ($S_{MB}$) scenarios. BBox importance also diminishes as pedestrians move farther away from the ego-vehicle ($S_{CP}$ vs. $S_{FP}$), reflecting diverse bounding box sizes due to variations in pedestrian distance. Furthermore, ego-vehicle acceleration (in $S_{Acc}$ scenario) introduces additional variability in BBox features due to rapid changes in relative positions.

    Pose importance scores may vary across different contexts as well. For instance, in the red-traffic light scenario ($S_{Red}$), importance scores are higher compared to $S_{Green}$ scenario. This could indicate that directional cues provided by pedestrians' poses become more informative when traffic conditions allow for crossing. The higher importance values for pose features in the stopped vehicle scenario ($S_{Stopped}$) suggest that pedestrian poses become more discernible and informative when vehicles are stationary.
    
    Speed importance scores change across scenarios with different ego-vehicle speeds. In the accelerating scenario ($S_{Acc}$), importance scores are higher compared to $S_{Dec}$ scenario, suggesting the model places more emphasis on vehicle speed when accelerating. Lower importance scores for decelerating imply the vehicle is more likely to yield to pedestrians, aiding correct crossing intention prediction. In $S_{Stopped}$ scenario, speed is always zero, thus shuffling the Speed feature does not affect models.

    The most variations in Local context feature importance scores are observed in scenarios with different proximity levels. In close proximity $S_{CP}$ scenario, it garners higher importance scores, perhaps due to providing richer environmental information.  
    Conversely, as pedestrian proximity decreases from medium ($S_{MP}$) to far ($S_{FP}$), the importance diminishes, potentially reducing its effectiveness in aiding model predictions.
    
\subsubsection{\textbf{Effects of Occlusion and Distance}}
    The Pose feature is the most unreliable, as it is influenced by the accuracy of the pose estimation algorithm. It tends to give inaccurate estimations, especially in samples with distant or occluded pedestrians. 
    The local context feature may also be impacted by occlusion caused by environmental obstacles such as parked vehicles and other road users. 
    However, Speed and BBox are the most accurate data points in the dataset. Even when occlusion occurs, BBox remains reliable and consistently captures the location of the pedestrian's full body.

\subsubsection{\textbf{Models Sensitivity}}
    The sensitivity is inferred by observing how the model's AUC changes in response to the permutation of different features. 
    SingleRNN demonstrates the least sensitivity to shuffling the BBox feature, with a decline in AUC of -6.6\% (${\sigma=1.4}$), compared to other models such as SFRNN (by -7.6\% with ${\sigma=1.8}$), PCPA (by -7.2\% with ${\sigma= 1.8}$), CAPformer (by -8.1\% with ${\sigma= 1.9}$), and VMIGI (by -8.5\% with ${\sigma=2.2}$). 
    
    VMIGI exhibits the highest sensitivity to the permutation of the Pose feature, with a decline in AUC of 1.3\% (${\sigma= 0.06}$), and compared to other models such as SFRNN (by -1\% with ${\sigma=0.2}$), PCPA (by 1.1\% with ${\sigma= 0.5}$), and CAPformer (by -1.1\% with ${\sigma= 0.4}$).

    CAPformer shows the highest sensitivity to the permutation of the Speed feature, with a decline in AUC of -5.2\% (${\sigma= 2.1}$), and compared to other models such as PCPA (by -4.5\% with ${\sigma= 1.9}$), and VMIGI (by -5\% with ${\sigma=2.2}$).

    VMIGI also shows the highest sensitivity to the permutation of the Local context feature, with a decline in AUC of -4.5\% (${\sigma= 1}$), and compared to other models such as SFRNN (by -3.5\% with ${\sigma=0.8}$), PCPA (by -4.3\% with ${\sigma= 1}$), and CAPformer (by -4\% with ${\sigma= 0.9}$).

\subsubsection{\textbf{Interaction Effects}}
    PCPA and VMIGI have also shown an importance score balancing behaviour in $S_{Stopped}$ scenario by increased importance for BBox (0.092 and 0.093) and Pose (0.018 and 0.034). The increased importance scores suggest that both models (PCPA and VMIGI) heavily rely on these spatial details to make informed decisions. This collaboration is facilitated by high-quality Pose information in this scenario and effective fusion strategies during training.

\subsection{Ego-Vehicle Motion Feature}
The assessment of cross-context permutation feature importance highlights the significant role of the speed parameter in predictive models. When focusing on scenarios where pedestrians intend to cross and the ego-vehicle's speed decreases ($S_{C} \cup S_{Dec}$), permuting the speed parameter with scenarios of constant ego-vehicle speed ($S_{Const}$) resulted in a notable decrease in prediction performance, with a -12.8\% decrease in AUC and a -9.7\% decrease in F1 score. Conversely, exchanging the speed parameter in scenarios where pedestrians don't intend to cross and the ego-vehicle speed is constant ($S_{NC} \cup S_{Const}$) with scenarios of decreasing speed ($S_{Dec}$) led to a reduction in prediction performance of -7.8\% in AUC and -10.3\% in F1 score.
These findings underscore a clear relationship between the speed parameter and predictive accuracy, suggesting that speed can introduce bias by capturing ego-vehicle behaviour rather than pedestrian behaviour.

The performance of the proposed motion feature representation, $\Delta_{P}$, as outlined in Section \ref{sec:motion}, is evaluated by training three intent-predictive models with default hyperparameters using the $\Delta_{P}$ feature instead of the speed parameter. This substitution aims to mitigate biased predictions influenced by speed. Table \ref{tab:delta} shows the performance of the models using the $\Delta_{P}$ feature with different $dt$ parameter values.

\begin{table}[htbp]
\centering
\caption{Performance Evaluation using $\Delta_{P}$ input}
\label{tab:delta}
\begin{tabular}{lcccc}
\noalign{\hrule height 1pt}
\multicolumn{1}{l}{\multirow{1}{*}{\textbf{Model}}} & \multicolumn{1}{c}{\textbf{$dt$}} & \multicolumn{1}{c}{\textbf{Acc}} & \multicolumn{1}{c}{\textbf{AUC}} & \multicolumn{1}{c}{\textbf{F1}} \\ \hline
\multicolumn{1}{l}{\multirow{3}{*}{SingleRNN}}    & 5  & 0.781 & 0.703 & 0.613 \\ 
                                                  & 10 & 0.802 & 0.729 & 0.676 \\ 
                                                  & 15 & 0.805 & 0.743 & 0.636 \\ \hline
\multicolumn{1}{l}{\multirow{3}{*}{SFRNN}}        & 5  & 0.779 & 0.731 & 0.652 \\ 
                                                  & 10 & 0.795 & 0.749 & 0.671 \\
                                                  & 15 & 0.809 & 0.763 & 0.679 \\ \hline
\multicolumn{1}{l}{\multirow{3}{*}{PCPA}}         & 5  & 0.791 & 0.773 & 0.712 \\ 
                                                  & 10 & 0.803 & 0.789 & 0.729 \\  
                                                  & 15 & 0.836 & 0.813 & 0.759 \\ 

\noalign{\hrule height 1pt}
\end{tabular}
\end{table}

The incorporation of the $\Delta_{P}$ feature shows no improvement in the models' performances. However, it appears to foster a more intricate understanding of the relationships between input features within the models. Repeating the cross-context permutation feature importance analysis for the $\Delta_{P}$ feature revealed a -6.9\% decrease in AUC and a -6.1\% decrease in F1 score in $S_{C} \cup S_{Dec}$ scenarios when permuting with $S_{Const}$ scenarios, and a -4.8\% decrease in AUC and an -8.3\% decrease in F1 score in $S_{NC} \cup S_{Const}$ scenarios when permuting with $S_{Dec}$ scenarios.

\section{Conclusion}

In this study, we conducted a comprehensive evaluation of five architecture-distinct intent-predictive models for pedestrian crossing scenarios using the Pedestrian Intention Estimation (PIE) dataset. Our experiments included context-aware performance evaluation, analysis of high-risk crossing scenarios, and assessment of input feature importance and ego-vehicle motion representations.
The performance evaluation revealed nuanced differences among candidate models across various contextual characteristics. Generally, models performed better in scenarios with decreasing ego-vehicle speed, designated crosswalks, and red traffic lights. However, midblock scenarios posed significant challenges, resulting in the lowest performance in cooperation with baseline performance.
Identifying high-risk crossing scenarios highlighted potential hazards if models fail to accurately detect pedestrian intentions, emphasising the importance of robust predictive capabilities, and the lack of large-scale datasets that capture a wide array of traffic contexts and edge-case scenarios.

Additionally, we not only evaluated the permutation feature importance across all contexts spread in the test and validation sets but also considered context-aware permutation feature importance by subdividing contexts. This approach enabled us to obtain more interpretable and reliable feature importance assessments with reduced variance in importance scores.

Feature importance analysis revealed the critical role of input features such as pedestrian bounding box, ego-vehicle speed, and local context features in predictive performance, and body pose is deemed less significant for models, potentially due to susceptibility to noise and occlusion. Despite variations in model architectures, there was a striking resemblance in how models responded to evaluations across various contexts and feature permutations, suggesting the fundamental relevance of certain features to the task.

Furthermore, our analysis of ego-vehicle motion features demonstrated the impact of speed on predictive accuracy, indicating potential biases introduced by capturing vehicle behaviour. While substituting the speed parameter with an implicit feature representation of ego-vehicle motions did not yield significant performance improvements, it provided insights into feature relationships within models by systematically evaluating feature importance across different contexts.

Overall, this study underscores the importance of considering contextual factors and diverse feature representations in developing accurate and robust intent-predictive models for pedestrian crossing scenarios. Future research should focus on addressing challenges in complex traffic environments, such as intersections with multiple turning lanes, and high pedestrian density areas (e.g., school zones and busy commercial districts). Additionally, exploring novel feature representations to enhance predictive capabilities and pedestrian safety is crucial, as this study was limited to the most common input features used in intent-predictive models, leaving many features yet to be assessed.

\section*{Declaration of competing interest}
The authors declare that they have no known competing financial interests or personal relationships that could have appeared to influence the work reported in this paper.

\section*{Acknowledgements}
The authors would like to thank all partners within the Hi-Drive project for their cooperation and valuable contribution. This research has received funding from the European Union’s Horizon 2020 research and innovation programme, under grant Agreement No 101006664. The article reflects only the author’s view and neither the European Commission nor CINEA is responsible for any use that may be made of the information this document contains.

\bibliographystyle{IEEEtran}
\bibliography{ref.bib}

\newpage

\section{Biography}

\vspace{-23pt}
\begin{IEEEbiography}[{\includegraphics[width=1in,height=1.25in,clip,keepaspectratio]{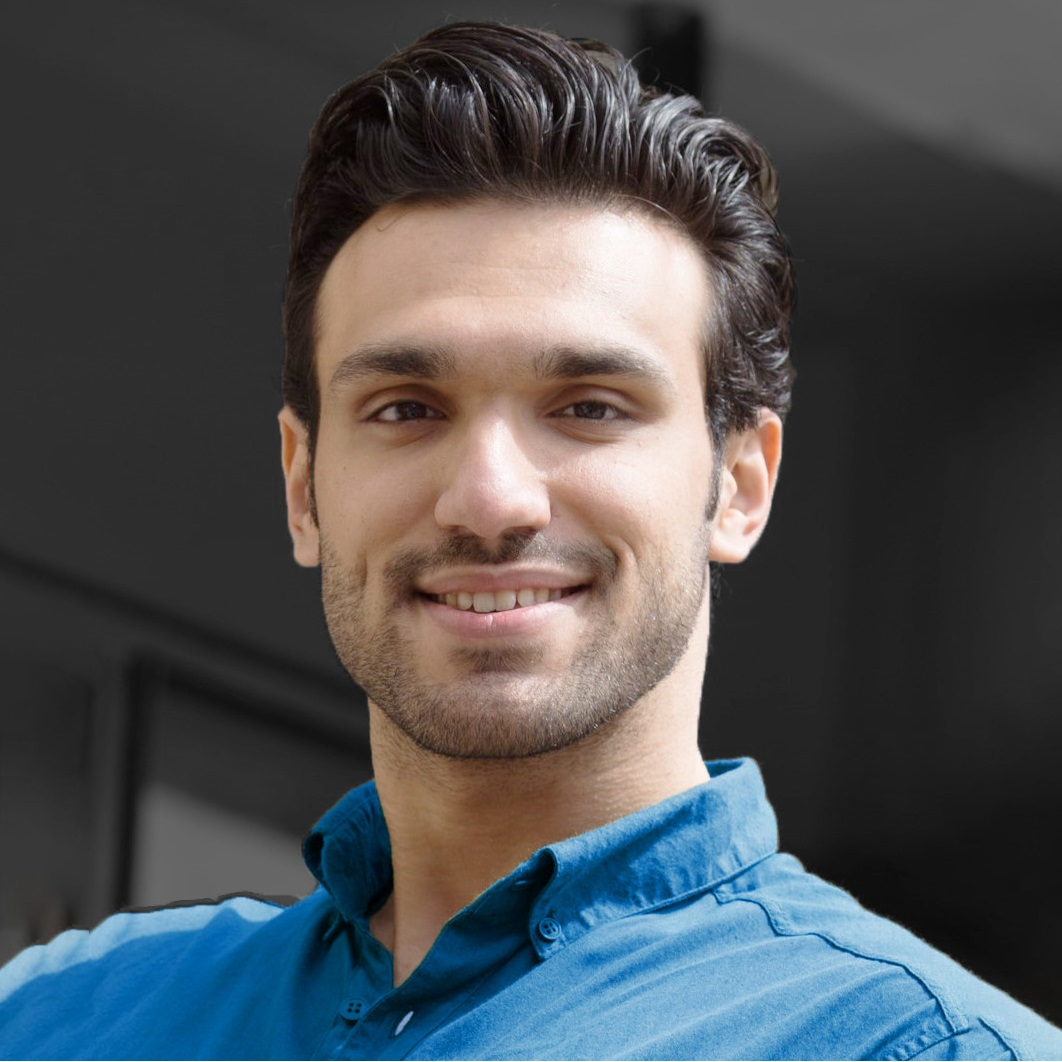}}]{Mohsen Azarmi}
is a Ph.D. Student at the University of Leeds, Institute for Transport Studies, UK.
He holds a master's degree in Artificial Intelligence \& Robotics and his main research direction and expertise are Computer Vision, Deep Neural Networks, and multi-sensor data fusion with a particular focus on pedestrian activity recognition, transportation and traffic safety. 
\end{IEEEbiography}

\vspace{-23pt}
\begin{IEEEbiography}[{\includegraphics[width=1in,height=1.25in,clip,keepaspectratio]{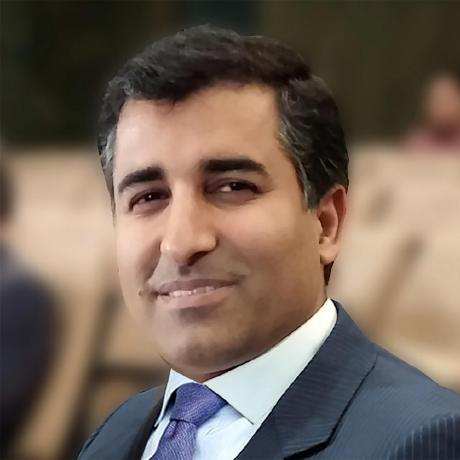}}]{Mahdi Rezaei}
is an Associate Professor of Computer Vision and Machine Learning and Leader of the Computer Vision Research Group at the University of Leeds, Institute for Transport Studies. He received his PhD in Computer Science from the University of Auckland, with the Top Doctoral Thesis Award in 2014. Offering 18 years of service and research experience in academia and industry, Dr Rezaei has published 60+ journals and conference papers in top-tier venues. He is also the Principal Investigator, lead Co-Investigator, or Collaborator of multiple European, UKRI, and EPSRC AV-related projects such as L3Pilot, Hi-Drive, IAA, Research England, and MAVIS.
\end{IEEEbiography}

\vspace{-23pt}
\begin{IEEEbiography}[{\includegraphics[width=1in,height=1.25in,clip,keepaspectratio]{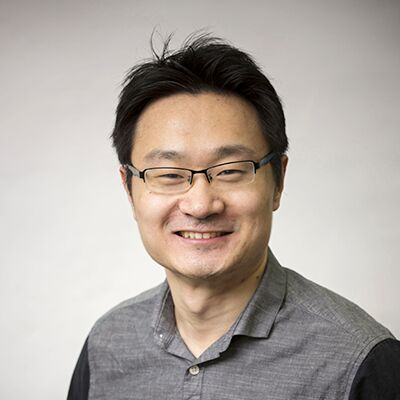}}]{He Wang}
is an Associate Professor at the Department of Computer Science, University College London (UCL) and a Visiting Professor at the University of Leeds. He is the Director of High-Performance Graphics and Game Engineering and Academic Lead of Centre for Immersive Technology. His current research interest is mainly in Computer Graphics, Vision and Machine Learning and applications. 
\end{IEEEbiography}

\vspace{-23pt}
\begin{IEEEbiography}[{\includegraphics[width=1in,height=1.25in,clip,keepaspectratio]{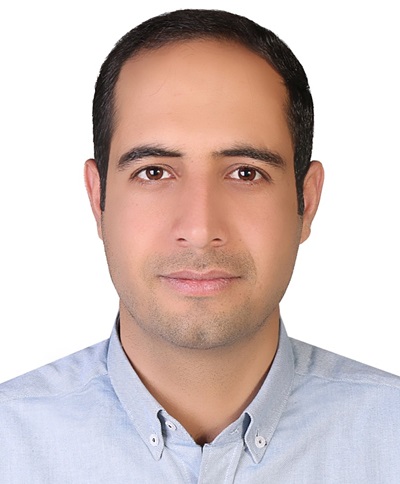}}]{Ali Arabian}
received the M.Sc. degree in ergonomics from the Tehran University of Medical Sciences, in 2019. He is currently pursuing a Ph.D. degree in transport studies at the Institute for Transport Studies, University of Leeds. His current research interest is mainly in the human factors of highly automated vehicles, particularly the allocation of visual attention during the transition from automated driving and HMI design.
\end{IEEEbiography}

\vspace{11pt}

\vfill

\newpage

\section*{Appendix}\label{sec:apdx}

This section presents detailed evaluation results of CAPFI scores for each model within specific contexts.

\begin{figure}[htbp]
    \centering
    \includegraphics[width=3.5in]{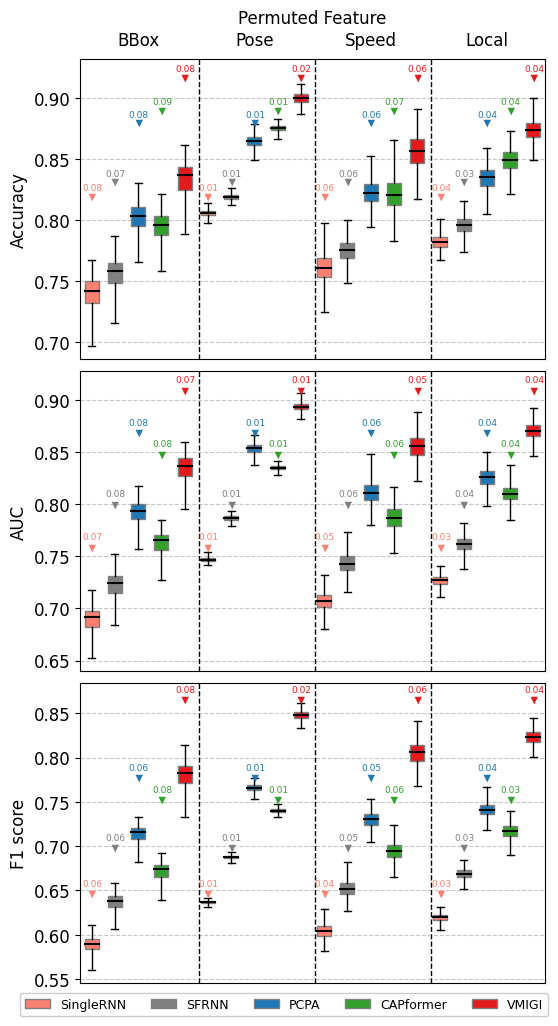}
    \caption{Scenario context: $S_{FW}$. \\ The number of samples = 441.}
    \label{fig:context_2_PFI}
\end{figure}

\begin{figure}[htbp]
    \centering
    \includegraphics[width=3.5in]{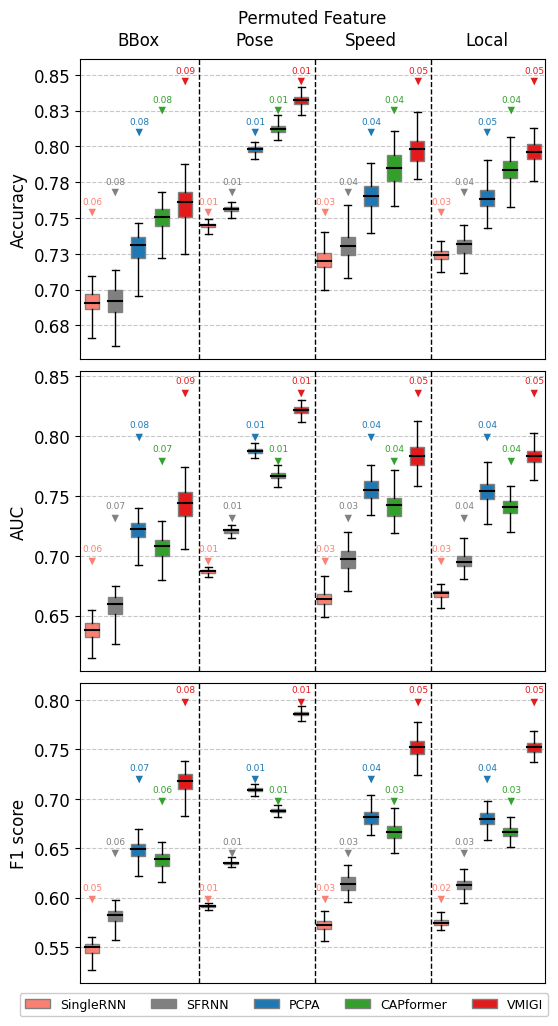}
    \caption{Scenario context: $S_{MB}$. \\ The number of samples = 164.}
    \label{fig:context_3_PFI}
\end{figure}

\begin{figure}[htbp]
    \centering
    \includegraphics[width=3.5in]{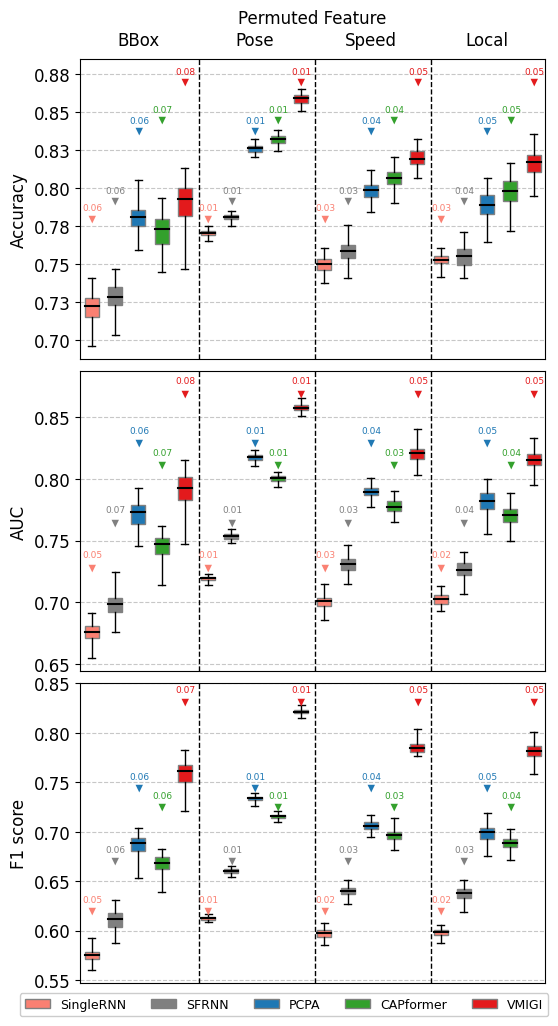}
    \caption{Scenario context: $S_{TJ}$. \\ The number of samples = 103.}
    \label{fig:context_4_PFI}
\end{figure}

\begin{figure}[htbp]
    \centering
    \includegraphics[width=3.5in]{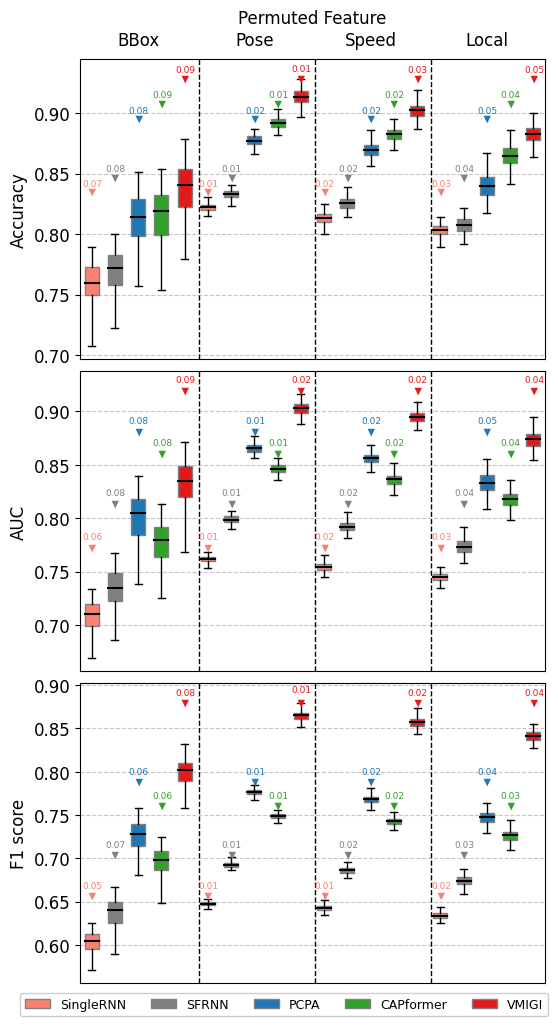}
    \caption{Scenario context: $S_{Red}$. \\ The number of samples = 93.}
    \label{fig:context_5_PFI}
\end{figure}

\begin{figure}[htbp]
    \centering
    \includegraphics[width=3.5in]{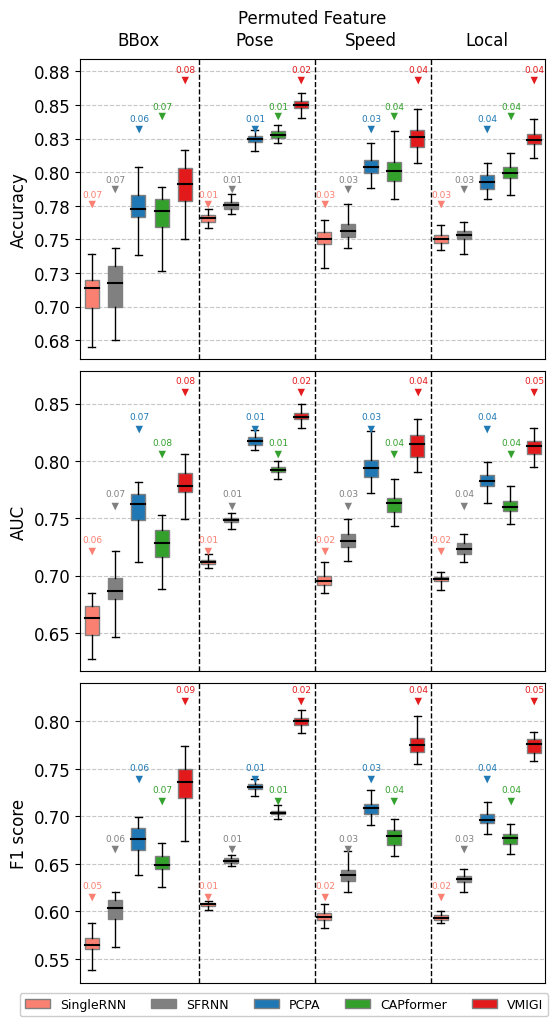}
    \caption{Scenario context: $S_{Yellow}$. \\ The number of samples = 37.}
    \label{fig:context_6_PFI}
\end{figure}

\begin{figure}[htbp]
    \centering
    \includegraphics[width=3.5in]{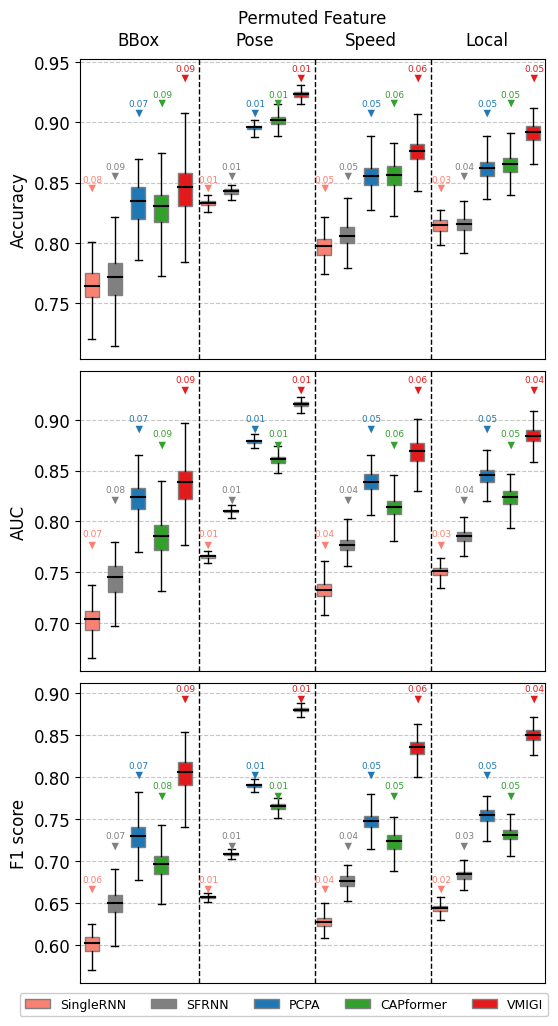}
    \caption{Scenario context: $S_{Green}$. \\ The number of samples = 242.}
    \label{fig:context_7_PFI}
\end{figure}

\begin{figure}[htbp]
    \centering
    \includegraphics[width=3.5in]{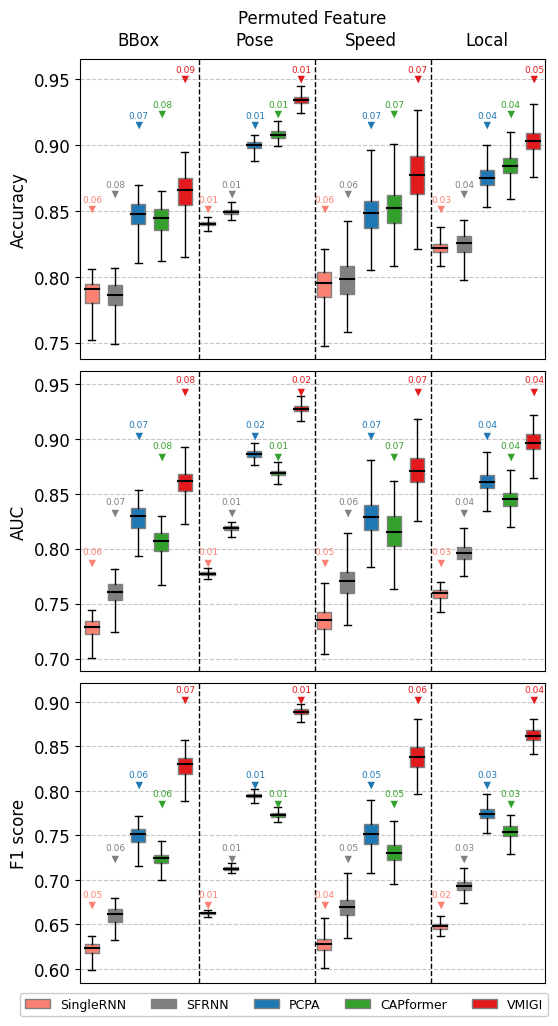}
    \caption{Scenario context: $S_{ZC}$. \\ The number of samples = 452.}
    \label{fig:context_8_PFI}
\end{figure}

\begin{figure}[htbp]
    \centering
    \includegraphics[width=3.5in]{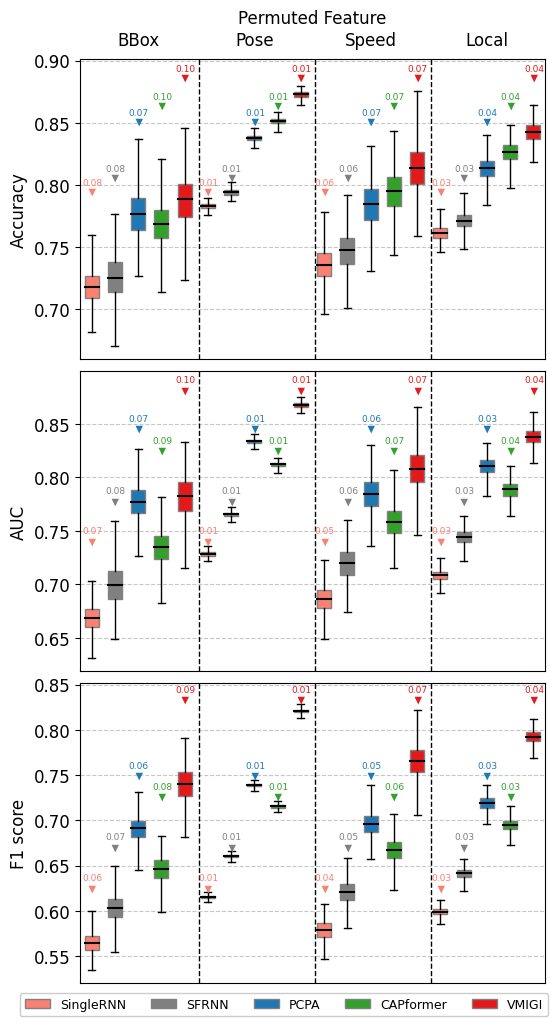}
    \caption{Scenario context: $S_{NZC}$. \\ The number of samples = 653.}
    \label{fig:context_9_PFI}
\end{figure}

\begin{figure}[htbp]
    \centering
    \includegraphics[width=3.5in]{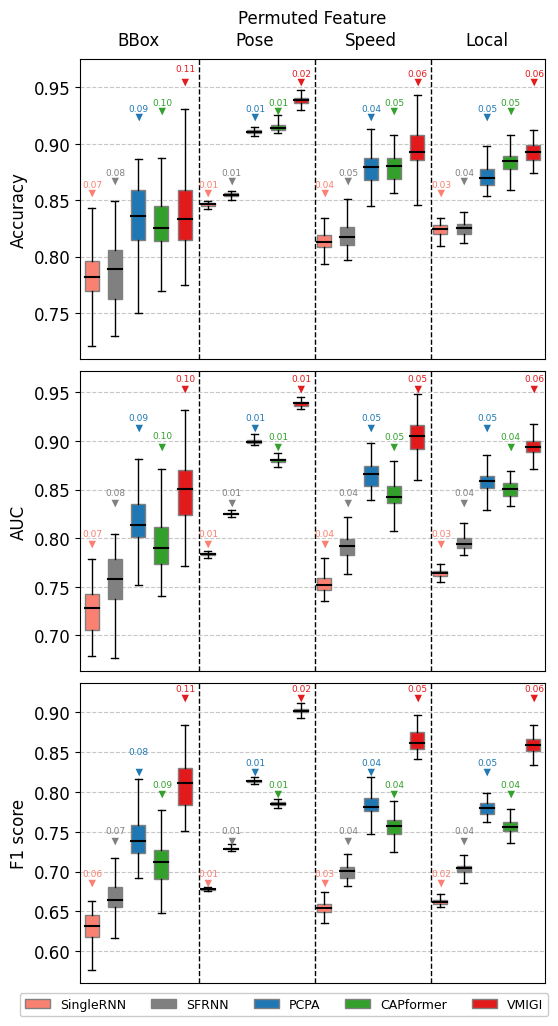}
    \caption{Scenario context: $S_{CP}$. \\ The number of samples = 58.}
    \label{fig:context_10_PFI}
\end{figure}

\begin{figure}[htbp]
    \centering
    \includegraphics[width=3.5in]{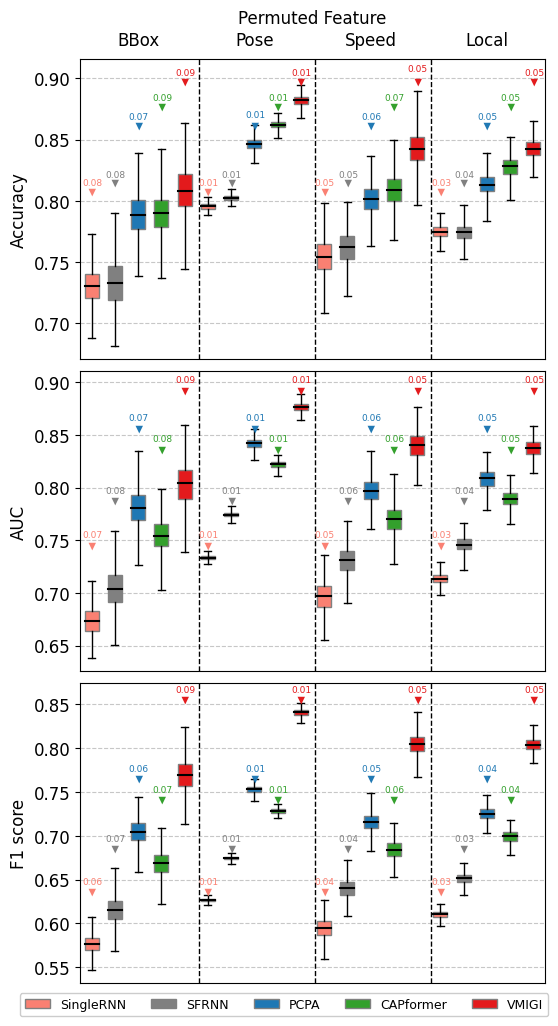}
    \caption{Scenario context: $S_{MP}$. \\ The number of samples = 542.}
    \label{fig:context_11_PFI}
\end{figure}

\begin{figure}[htbp]
    \centering
    \includegraphics[width=3.5in]{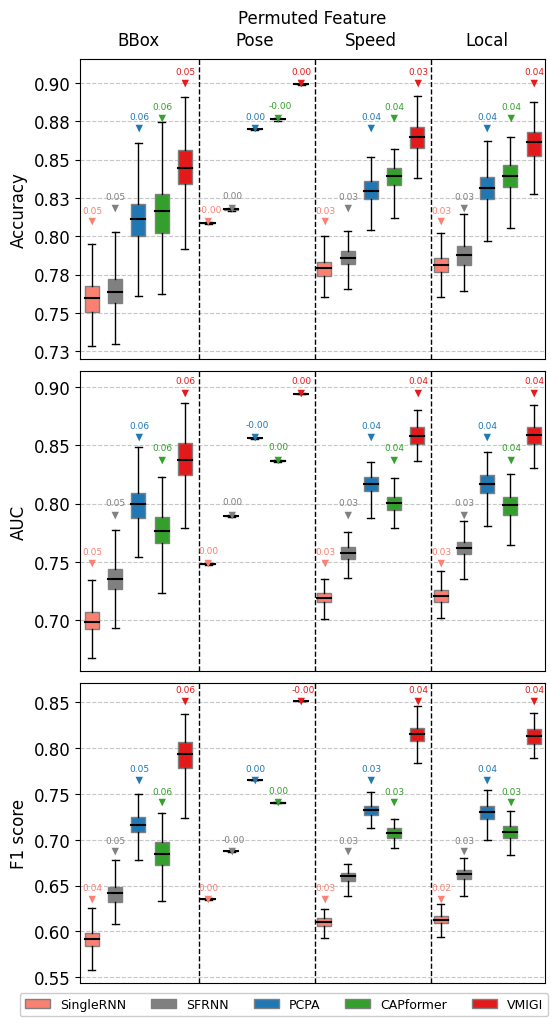}
    \caption{Scenario context: $S_{FP}$. \\ The number of samples = 291.}
    \label{fig:context_12_PFI}
\end{figure}

\begin{figure}[htbp]
    \centering
    \includegraphics[width=3.5in]{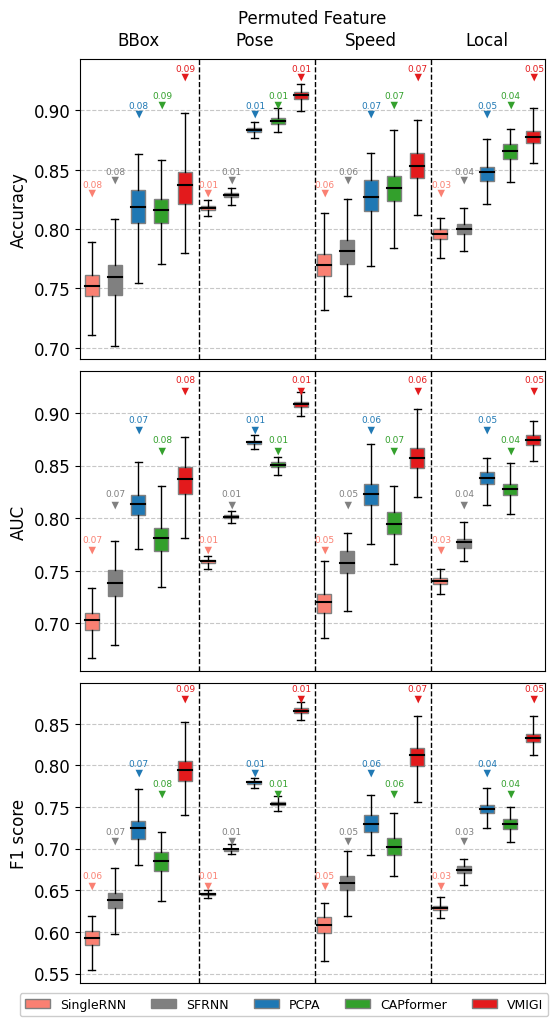}
    \caption{Scenario context: $S_{Acc}$. \\ The number of samples = 216.}
    \label{fig:context_13_PFI}
\end{figure}

\begin{figure}[htbp]
    \centering
    \includegraphics[width=3.5in]{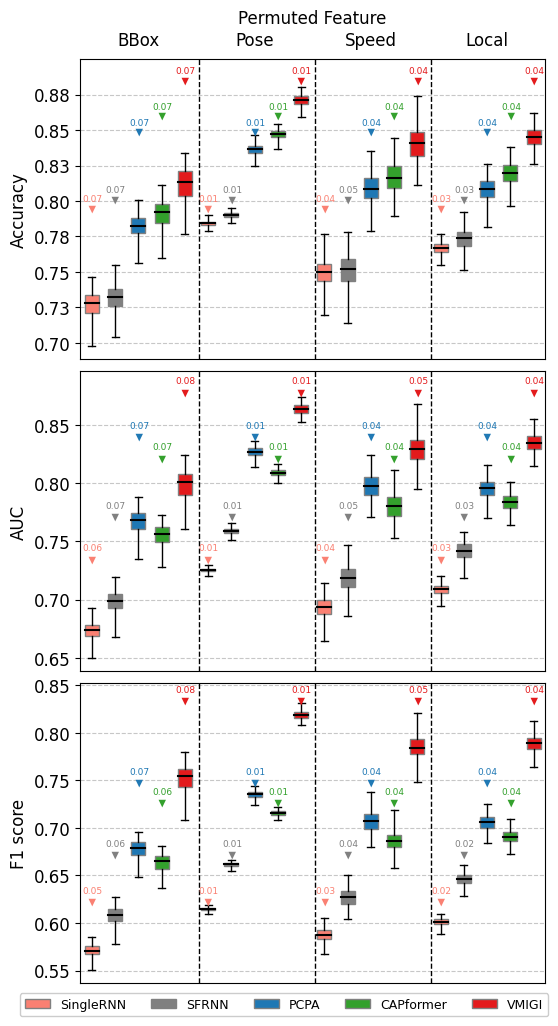}
    \caption{Scenario context: $S_{Const}$. \\ The number of samples = 298.}
    \label{fig:context_14_PFI}
\end{figure}

\begin{figure}[htbp]
    \centering
    \includegraphics[width=3.5in]{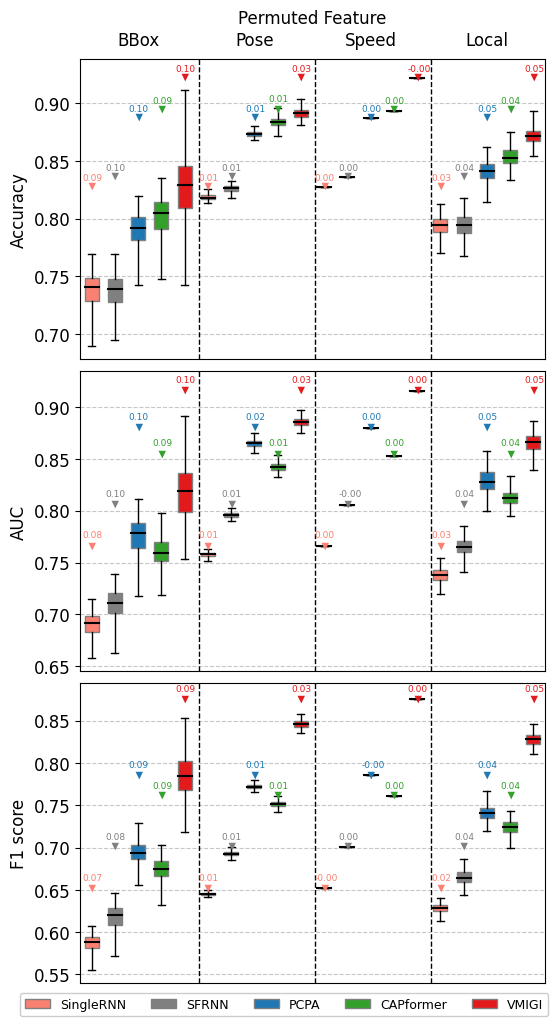}
    \caption{Scenario context: $S_{Stopped}$. \\ The number of samples = 185.}
    \label{fig:context_15_PFI}
\end{figure}

\begin{figure}[htbp]
    \centering
    \includegraphics[width=3.5in]{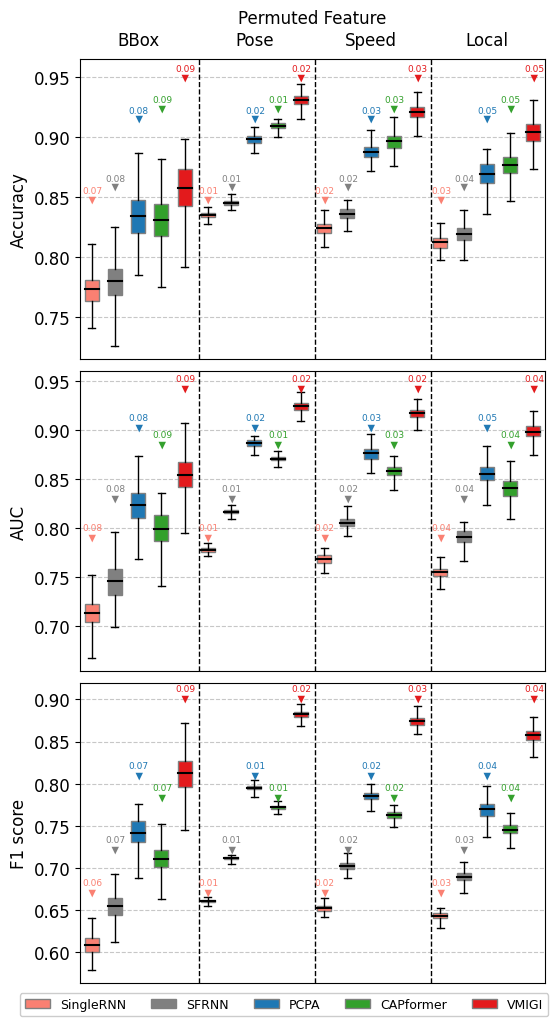}
    \caption{Scenario context: $S_{Dec}$. \\ The number of samples = 193.}
    \label{fig:context_16_PFI}
\end{figure}

\end{document}